\documentclass{article}

\usepackage{microtype}
\usepackage{graphicx}
\usepackage{subfigure}
\usepackage{booktabs} % for professional tables

\usepackage{graphicx}
\usepackage{amsmath}
\usepackage{amssymb}
\usepackage{multirow}
\usepackage{arydshln}
\usepackage{comment}
\usepackage{enumitem}
\usepackage[table,x11names]{xcolor}

\usepackage{hyperref}

\usepackage[accepted]{icml2024}
\usepackage{amsmath}
\usepackage{amssymb}
\usepackage{mathtools}
\usepackage{amsthm}

\usepackage[capitalize,noabbrev]{cleveref}

\theoremstyle{plain}

\theoremstyle{definition}

\theoremstyle{remark}

\usepackage[textsize=tiny]{todonotes}

\def\eg{{\em e.g.}}

\newcommand{\myPara}[1]{\vspace{.01in}\noindent\textbf{#1}}

\newcommand{\figref}[1]{Fig. \ref{#1}}

\newcommand{\bl}[1]{\textbf{#1}}

\newcommand{\mc}[1]{\mathcal{#1}}
\newcommand{\mb}[1]{\mathbb{#1}}

\graphicspath{{./fig/}}

\icmltitlerunning{Masked Face Recognition with Generative-to-Discriminative Representations}

\begin{document}

\twocolumn[
\icmltitle{Masked Face Recognition with Generative-to-Discriminative Representations}

% It is OKAY to include author information, even for blind
% submissions: the style file will automatically remove it for you
% unless you've provided the [accepted] option to the icml2024
% package.

% List of affiliations: The first argument should be a (short)
% identifier you will use later to specify author affiliations
% Academic affiliations should list Department, University, City, Region, Country
% Industry affiliations should list Company, City, Region, Country

% You can specify symbols, otherwise they are numbered in order.
% Ideally, you should not use this facility. Affiliations will be numbered
% in order of appearance and this is the preferred way.
\icmlsetsymbol{equal}{*}

\begin{icmlauthorlist}
\icmlauthor{Shiming Ge}{1,2}
\icmlauthor{Weijia Guo}{1,2}
\icmlauthor{Chenyu Li}{1,2,3}
\icmlauthor{Junzheng Zhang}{1,2}
\icmlauthor{Yong Li}{1,2}
\icmlauthor{Dan Zeng}{4}
\end{icmlauthorlist}
\icmlaffiliation{1}{Institute of Information Engineering, Chinese Academy of Sciences, Beijing 100092, China.}
\icmlaffiliation{2}{School of Cyber Security at University of Chinese Academy of Sciences, Beijing 100049, China.}
\icmlaffiliation{3}{Cloud Music Inc., Hangzhou 311215, China.}
\icmlaffiliation{4}{Department of Communication Engineering, Shanghai University, Shanghai 200040, China}

\icmlcorrespondingauthor

% You may provide any keywords that you
% find helpful for describing your paper; these are used to populate
% the "keywords" metadata in the PDF but will not be shown in the document
\icmlkeywords{Machine Learning, ICML}

\vskip 0.3in
]

% this must go after the closing bracket ] following \twocolumn[ ...

% This command actually creates the footnote in the first column
% listing the affiliations and the copyright notice.
% The command takes one argument, which is text to display at the start of the footnote.
% The \icmlEqualContribution command is standard text for equal contribution.
% Remove it (just {}) if you do not need this facility.

\printAffiliationsAndNotice{}  % leave blank if no need to mention equal contribution
%\printAffiliationsAndNotice{\icmlEqualContribution} % otherwise use the standard text.

\begin{abstract}
Masked face recognition is important for social good but challenged by diverse occlusions that cause insufficient or inaccurate representations. In this work, we propose a unified deep network to learn generative-to-discriminative representations for facilitating masked face recognition. To this end, we split the network into three modules and learn them on synthetic masked faces in a greedy module-wise pretraining manner. First, we leverage a generative encoder pretrained for face inpainting and finetune it to represent masked faces into category-aware descriptors. Attribute to the generative encoder's ability in recovering context information, the resulting descriptors can provide occlusion-robust representations for masked faces, mitigating the effect of diverse masks. Then, we incorporate a multi-layer convolutional network as a discriminative reformer and learn it to convert the category-aware descriptors into identity-aware vectors, where the learning is effectively supervised by distilling relation knowledge from off-the-shelf face recognition model. In this way, the discriminative reformer together with the generative encoder serves as the pretrained backbone, providing general and discriminative representations towards masked faces. Finally, we cascade one fully-connected layer following by one softmax layer into a feature classifier and finetune it to identify the reformed identity-aware vectors. Extensive experiments on synthetic and realistic datasets demonstrate the effectiveness of our approach in recognizing masked faces.
\end{abstract}

\section{Introduction}\label{sec:intro}

\begin{figure}[!t]
  \centering
\includegraphics[width=1.0\linewidth]{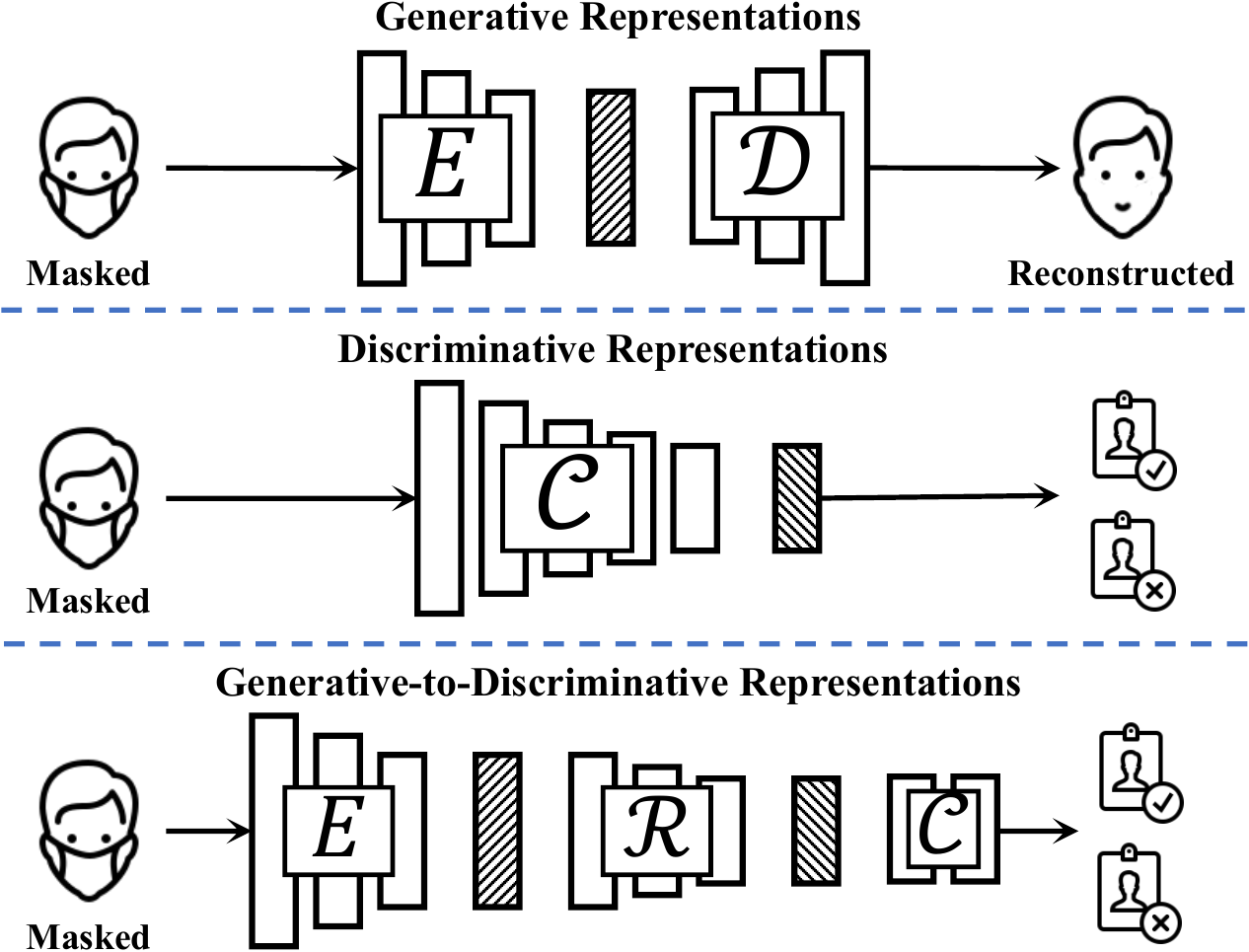}
  \caption{Our approach learns generative-to-discriminative representations for masked face recognition, which combines the advantages of generative representations and discriminative representations, providing general and robust solution to recover missing clues and capture identity-related characteristics.}
  \label{fig:motivation}
\end{figure}

Deep face recognition models have delivered impressive performance on public benchmarks~\cite{wen2016discriminative,cao2018vggface2,deng2019arcface} and realistic scenarios~\cite{anwar2020masked}. In general, these models are designed for recognizing unmasked faces and often suffer from sharp accuracy drop in recognizing masked faces~\cite{ngan2020a,ngan2020b}, which hinders real-world applications~\cite{ge2017detecting,delphine2022tip,zhang2023tmm,wang2023tbbis,alnabulsi2023sensors}. Unlike normal face recognition, masked face recognition is challenged by \emph{insufficient or inaccurate representations}. Masks often occlude some important facial features, causing key information loss. With the occluded regions growing larger, the unmasked regions may become less sufficient for accurate identity prediction. Moreover, the ill-posed mapping between observation and possible groundtruth faces makes representations inaccurate. Therefore, an effective solution for masked face recognition should learn representations that could recover missing facial clues and calibrate inaccurate identity clues. Accordingly, many masked face recognition approaches have been proposed, which are based on generative or discriminative idea~\cite{deng2021cvprw}.

``Generative'' approaches aim to reconstruct the missing facial clues then perform recognition on completed faces.  Deep generative models~\cite{zheng2023icml,choi2023icml} can provide general representations to noisy images~\cite{chen2020icml,he2022masked,li2023mage}. Generative face inpainting~\cite{pathak2016context,li2017generative,yu2018generative,zhao2018tip,yu2019iccv,wan2021high,dey2022cvpr,yang2023tmm} has successfully enabled recovery of high-quality visual contents. While it is very robust to generate consistent results to different masks, the benefit of face inpainting for masked face recognition is limited~\cite{joe2019icb} since they ignore the regularization on identity preservation, and the information about intra-identity and inter-identity relationships are lost during the process. Some attempts are made to enforce identity preservation via introducing face recognizer for regularization~\cite{zhang2017demeshnet,ge2020tcsvt}, yet the help is limited, since the recognizers in these approaches cannot provide the upstream inpainting network with direct feedback and enough regularization on identity awareness.

In contrast, ``discriminative'' approaches aim to extract robust representations by reducing the effect of masked regions~\cite{wen2016discriminative,he2022tip,song2019iccv}. They usually adopt part detection~\cite{ding2020masked}, compositional models~\cite{kortylewski2021compositional}, knowledge transfer~\cite{huber2021fg,boutros2022self,zhao2022spl} or complementary attention~\cite{cho2023fg}, to localize or remove masked regions. However, the unmasked regions often hardly provide enough information for accurate recognition. Thus, some approaches propose to finetune existing face recognizers~\cite{neto2021focusface} or design powerful networks~\cite{qiu2021end2end,boutros2022self,zhu2023tifs,zhao2024tift} to extract more information. Generally, directly finetuning general face recognizers may increase the accuracy on masked faces, while on the sacrifice of discriminative and generalization ability on the recognition of unmasked faces. Moreover, since the occlusions contain diverse mask types, these approaches usually show poor robustness on masked faces. In real-world masked face recognition scenarios, diverse masks could cause semantic divergences while the representations are expected to be consistent. Therefore, a key issue that needs to be carefully addressed in masked face recognition is the coordination between reconstructing general representations and enhancing their identity discrimination.

To facilitate masked face recognition, we propose learning generative-to-discriminative representations which combines the advantages from generative and dicriminative representations (Fig.~\ref{fig:motivation}). Specially, we cascade three modules and learn them in a greedy manner. First, generative encoder takes the encoder of a pretrained face inpainting model, and represents masked faces into category-aware descriptors with rich general information of masked faces to distinguish human faces from other objects. Then, discriminative reformer incorporates a 22-layer convolutional network and is learned to convert the category-aware descriptors into identity-aware vectors for enhancing recognition. Finally, feature classifier cascades a fully-connected layer and a softmax layer to identify the reformed vectors. In the approach, generative encoder and discriminative reformer are combined together, which serves as backbone for masked facial feature extraction and is progressively pretrained in a self-supervised manner, while feature classifier serves as recognition head. Finally, the backbone is frozen and feature classifier is finetuned on labeled masked faces.

Our main contributions can be summarized as: 1) we propose to learn generative-to-discriminative representations for masked face recognition, which combines the advantages of generative and discriminative representations to extract general and discriminative features for identifying masked faces; 2) we cascade generative encoder and discriminative reformer as the backbone and present a greedy module-wise pretraining strategy to improve representation learning via distillation in a self-supervised manner; and 3) we conduct extensive experiments and comparisons to demonstrate the effectiveness of our approach.

\section{Approach}

\begin{figure*}[!t]
  \centering
  \includegraphics[width=1.0\linewidth]{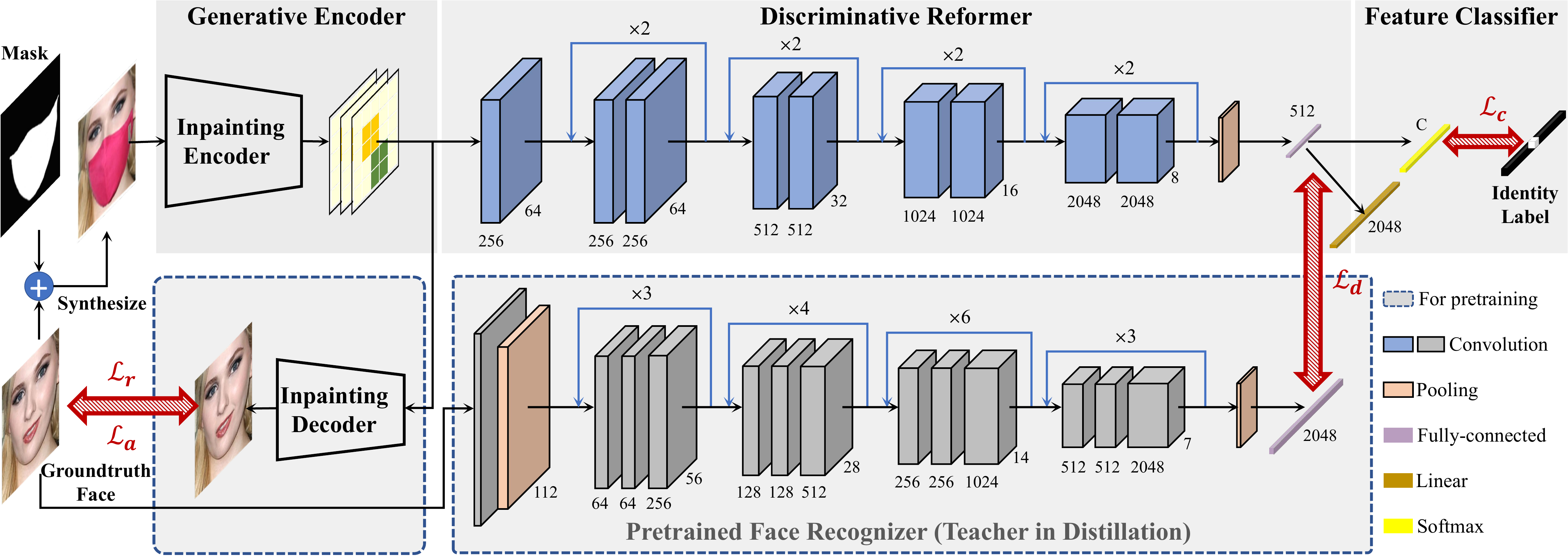}
  \caption{The framework of the proposed approach. It cascades three modules into a unified network and learns generative-to-discriminative representations on synthetic masked faces in a progressive manner. The approach first finetunes a generative encoder to represent a masked face into category-aware descriptors by initializing with a pretrained face inpainting model and finetuning via self-supervised pixel reconstruction. Then, it learns a CNN-based discriminative reformer to convert the category-aware descriptors into an identity-aware vector by distilling a general pretrained face recognizer via self-supervised relation-based feature approximation. Finally, it learns a feature classifier on identity-aware vectors by optimizing supervised classification task.}
  \label{fig:architecture}
\end{figure*}

\subsection{Problem Formulation}
Our objective is learning a deep model $\phi(\bl{x},\bl{m};\bl{w})$ for discriminative representations to identify a masked face $\bl{x}$. Here, the binary map $\bl{m}$ indicates whether a pixel $p$ is occluded ($\bl{m}(p)=1$) or not ($\bl{m}(p)=0$), and $\bl{w}$ are model parameters. Let $\hat{\bl{x}}$ denote the groundtruth but generally unavailable unmasked face, having $\bl{x}(p)=\hat{\bl{x}}(p)$ when $\bl{m}(p)=0$. Unlike the recognition of $\hat{\bl{x}}$, masked face recognition needs to learn representations from partly occluded faces where some informative clues are missing. Thus the key is to address the recovery of the missing clues from $\bl{x}$ to approximate the latent representations of $\hat{\bl{x}}$, ideally having:
\begin{equation}
\begin{aligned}
  \phi(\bl{x},\bl{m};\bl{w}) \doteq \Phi(\hat{\bl{x}};\hat{\bl{w}}),
\end{aligned}
\label{eq:formulation}
\end{equation}
where $\Phi(\cdot)$ is a deep face recognizer well-trained on unmasked faces with parameters $\hat{\bl{w}}$. The symbol $\doteq$ means ``equivalence'' in some metric (\eg, similarity of representations or consistency of predictions). To solve Eq.~\eqref{eq:formulation}, there are {three main challenges}: 1) \emph{greater complexity} due to the consideration of the joint distribution of $\bl{x}$ and $\bl{m}$, 2) \emph{consistency requirement} that expects to extract consistent representations even when masked faces originated from the same $\hat{\bl{x}}$ have diverse $\bl{m}$, and 3) \emph{insufficient data} due to difficulty of collecting real-world pairs $\{\hat{\bl{x}},\bl{x}\}$. To sum up, an effective solution for modeling masked faces is to learn representations through recovery, solving information reconstruction and representation clustering regularization in a unified and implicit way on synthetic data. 

As shown in Fig.~\ref{fig:architecture}, we address masked face recognition by learning generative-to-discriminative representations. The unified network $\phi$ consists of generative encoder $\phi_e$, discriminative reformer $\phi_r$ and feature classifier $\phi_c$ with parameters $\bl{w}_e$, $\bl{w}_r$ and $\bl{w}_c$, respectively. Given synthesized triplets $\{\hat{\bl{x}}_i,\bl{m}_i,\bl{x}_i\}_{i=1}^{n}$ with $n$ samples, $\phi_e$ and $\phi_r$ are first learned by progressively reconstructing appearance and latent features, which are solved by minimizing the appearance reconstruction loss $\mc{L}_{r}$ and latent loss $\mc{L}_{d}$, separately:
\begin{equation}
  \mc{L}_{r}(\bl{w}_e,\bl{w}_d)=\sum_{i=1}^{n}\ell(\psi(\phi(\bl{x}_i,\bl{m}_i;\bl{w}_e);\bl{w}_d),\hat{\bl{x}}_i),
\label{eq:rec}
\end{equation}
\begin{equation}
  \mc{L}_{d}(\bl{w}_r) = \sum_{i=1}^{n}\ell(\phi(\bl{x}_i,\bl{m}_i;\{\bl{w}_e,\bl{w}_r\}),\Phi(\hat{\bl{x}};\hat{\bl{w}})),
\label{eq:rel}
\end{equation}
where $\psi(\cdot)$ is inpainting decoder for training the encoder parameters $\bl{w}_e$ only. $\bl{w}_d$ are the decoder parameters. The trained $\bl{w}_e$ are then fixed and used in training the reformer parameters $\bl{w}_r$ by minimizing $\mc{L}_{d}$. $\Phi(\cdot)$ is a pretrained face recognizer used to guide the feature reconstruction of $\phi_r$, with $\hat{\bl{w}}$ as its model parameters. $\ell(\cdot)$ denotes the distance function. Finally, $\bl{w}_e$ and $\bl{w}_r$ are frozen and all three modules are cascaded for finetuning classification loss $\mc{L}_c$ to learn the feature classifier in a supervised way:
\begin{equation}
  \mc{L}_{c}(\bl{w}_c)=\sum_{i=1}^{n}\ell(\phi(\bl{x}_i,\bl{m}_i;\{\bl{w}_e,\bl{w}_r,\bl{w}_c\}),c_i),
\label{eq:cls}
\end{equation}
where $c_i$ denotes the groundtruth identity label for $\bl{x}_i$.

This architecture design can well address the {three main challenges} mentioned above in masked face recognition. First, the generative encoder and discriminative reformer are cascaded for the backbone, which decouples the burden of modeling \emph{greater complexity} by jointly handling information reconstruction and representation clustering regularization in a progressive way. Second, the encoder aims to output a consistent reconstruction for the given masked faces originated from a same unmasked face regardless of diverse masks, that meets the \emph{consistency requirement} of extracted representations. Third, it is easy to train the backbone on synthetic data in a self-supervised manner, alleviating the issue of \emph{insufficient data} to avoid expensive and time-consuming annotation of training samples.

\subsection{Generative Encoder}
Generative encoder is responsible for extracting {general} face representations under mask occlusion. It is derived from ICT~\cite{wan2021high} pretrained on FFHQ~\cite{karras2019cvpr}, one of the state-of-the-art Transformer-based inpainting method. It consists of a Transformer network for face representations and a CNN for upsampling faces. We extract generative representations from the middle residual block of the upsample network. Given an input image and a binary mask of size $256\times256$, the encoder computes a $64\times64\times256$ generative representation. To better adapt to the synthetic masked faces, we fix the Transformer and finetune the generative encoder on our training data with pixel reconstruction loss $\mc{L}_{r}$ as well as adversarial loss $\mc{L}_{a}$:
\begin{equation}
\begin{aligned}
  \mc{L}_{r}(\bl{w}_e,\bl{w}_r)&=\sum_{i=1}^n\|\tilde{\bl{x}}_i-\hat{\bl{x}_i}\|^2,\\
  \mc{L}_{a}=E_{\hat{\bl{x}}~\mb{R}}[\zeta(\hat{\bl{x}}_i)]-E_{\hat{\bl{x}}~\mb{G}}&[\zeta(\tilde{\bl{x}}_i)]+ E_{\check{\bl{x}}}(\|\nabla_{\check{\bl{x}}}\zeta(\check{\bl{x}})\|^2),
\end{aligned}
\label{eq:Lrec}
\end{equation}
where adversarial loss is defined using modified WGAN-GP~\cite{gulrajani2017improved}, $\mb{R}$ is the real face distribution, $\mb{G}$ is the distribution implicitly defined by $\psi(\phi(*))$,  $\Tilde{\bl{x}}=\psi(\phi(\bl{x}_i,\bl{m}_i;\bl{w}_e);\bl{w}_d)$ is the inpainted face, $\zeta$ denotes the discriminators, $\check{\bl{x}}$ is sampled from the straight line between $\mb{G}$ and $\mb{R}$, having $\nabla_{\check{\bl{x}}}\zeta(\check{\bl{x}})={(\tilde{x}-\check{\bl{x}})}/{\|\tilde{x}-\check{\bl{x}}\|}$.

We visualize the learned generative representations to check the consistency over diverse masks and clustering behaviors 
with t-SNE~\cite{maaten2008visualizing} in \figref{fig:representations}, finding high-overlapping among the same groundtrue faces and scatters among  different groundtrue faces. It implies that the generative representations can eliminate mask effect and are robust towards diverse masks, but can not well describe inter- and intra-identity characteristics. 

\subsection{Discriminative Reformer}
Discriminative reformer aims to turn the encoded generative representations into discriminative representations, so that the identity attributes can be better recovered and described. We cascade encoder and reformer as the backbone, which has several advantages. First, it reduces the accumulation of deviations. The reformer can shift the mapping from image space to latent space, avoiding the re-mapping loss during encoding of the completed faces. Second, latent space of higher level in neural network is proved to have flatter landscape~\cite{bengio2013better}, so the reformation in latent manifold is more understandable for face representations. Third, it can make better use of the information that high-level representations contains, such as long-distance dependence. Finally, feature reformation can be seamlessly integrated with the recognition head, allowing more efficient end-to-end optimization. We apply a Resnet-like network due to its effectiveness in face representation~\cite{cao2018vggface2,deng2019arcface} to construct the reformer, which consists of a convolutional layer, 4 residual blocks following by a pooling and a fully-connected layers, outputs 512$d$ vectors, as shown in~\figref{fig:architecture}. We have experimentally found that shallower structures are poor in converting generative representations into discriminative ones, while deeper or Transformer-based networks are effective but greatly increase model complexity.

\begin{figure}[!t]
  \centering
  \includegraphics[width=0.49\linewidth]{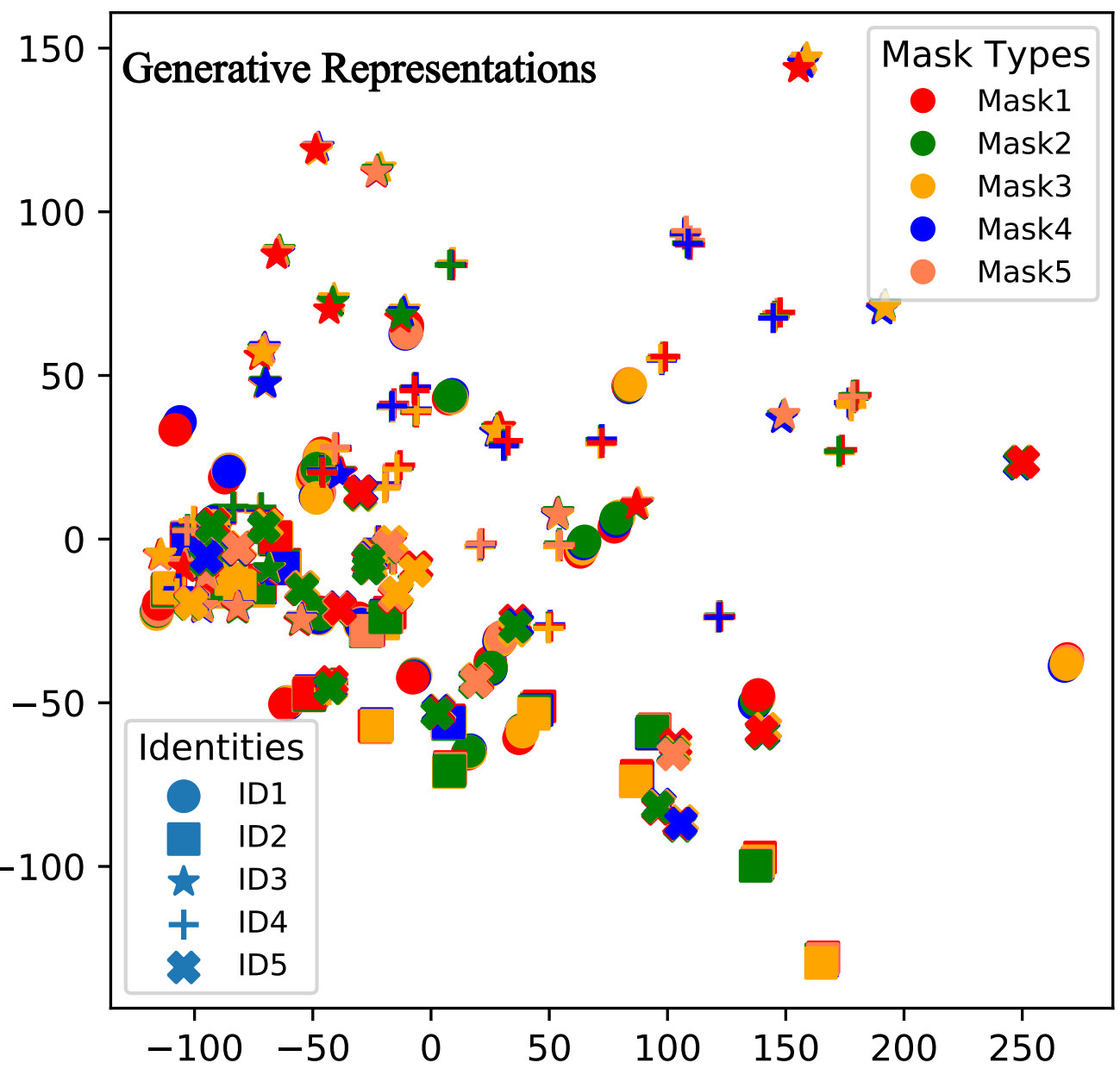}
  \includegraphics[width=0.49\linewidth]{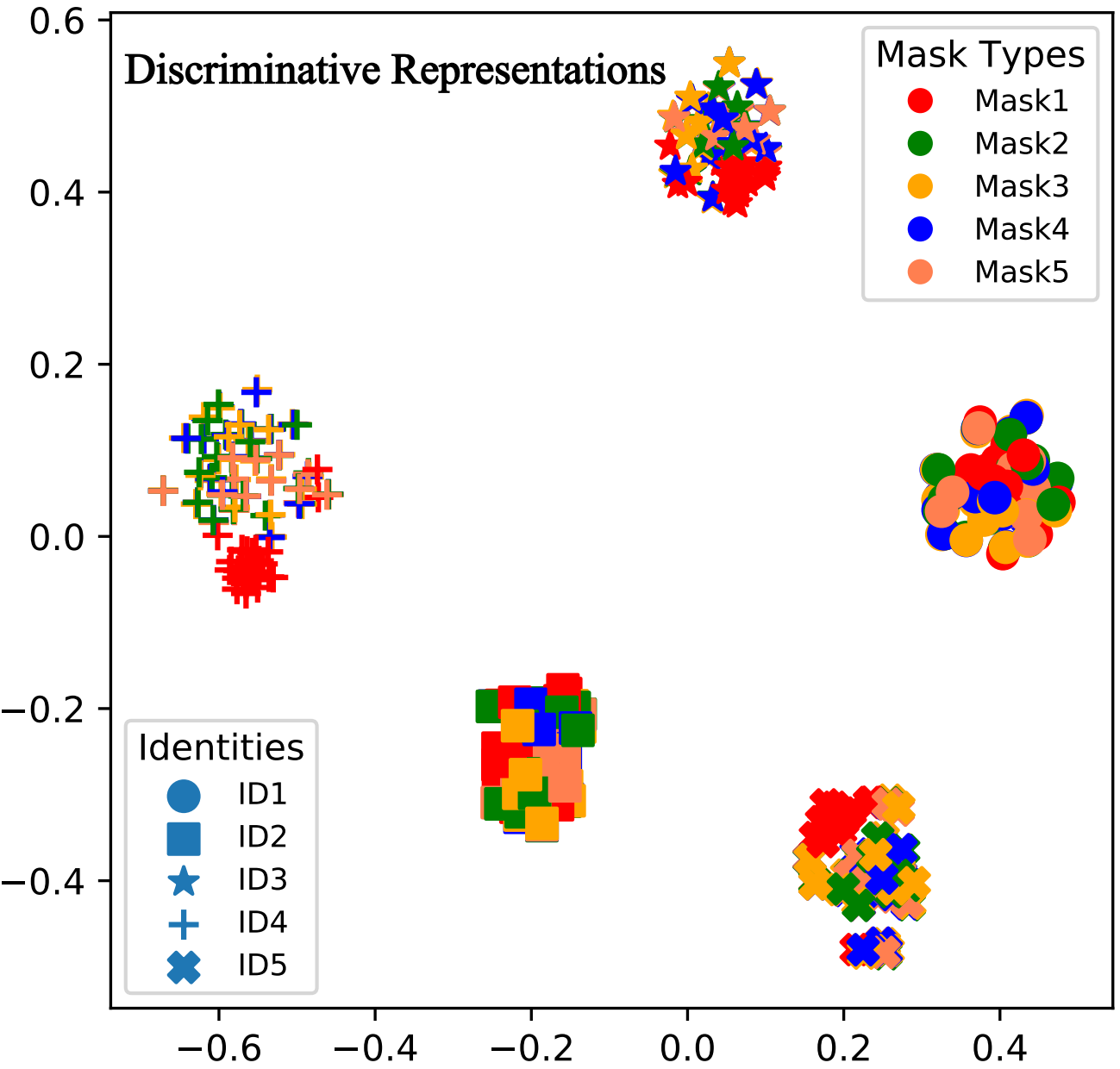}
  \includegraphics[width=0.98\linewidth]{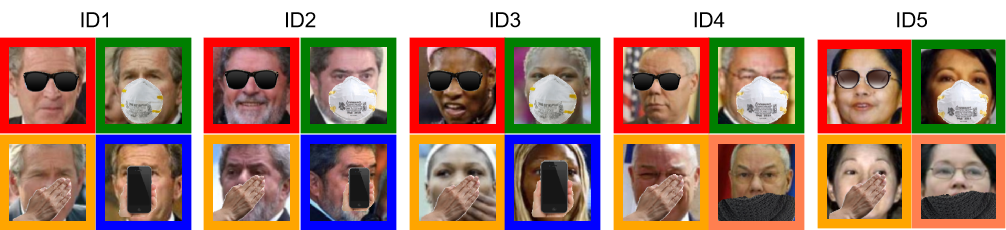}
  \caption{The t-SNE visualization of representations. We randomly sample five identities, use all sample images with these identities to synthesize masked faces with five random mask types, and extract generative and discriminative representations of masked faces. Generative representations are robust towards diverse mask occlusions but short in inter- and intra-identity discriminablility, while discriminative representations show good identity discriminablility. Bottom: some synthetic masked faces.}
  \label{fig:representations}
\end{figure}

Inspired by previous success in integrating external knowledge to facilitate optimization of neural networks~\cite{hinton2015distilling,wonpyo2019rkd,li2020mm}, we take a pretrained general face recognizer as teacher to guide the generative-to-discriminative representation reforming via knowledge distillation, and leverage essential guidance from unmasked faces for reforming and represent the teacher knowledge with two and three order structural relations:
\begin{equation}\label{eq:L_R2}
\begin{aligned}
  \mathcal{L}_{2} = \sum_{(i,j)\in \bl{S}_2} \ell_H(\frac{1}{\mu_\bl{t}} \|\bl{t}_i - &\bl{t}_j\|^2, \frac{1}{\mu_\bl{s}} \|\bl{s}_i - \bl{s}_j\|^2),
\end{aligned}
\end{equation}
\begin{equation}\label{eq:L_R3}
\begin{aligned}
  \mathcal{L}_{3}=\sum_{(i,j,k)\in\bl{S}_3}\ell_H(&\langle\frac{\textbf{t}_i-\textbf{t}_j}{\|\textbf{t}_i-\textbf{t}_j\|^2},\frac{\textbf{t}_k-\textbf{t}_j}{\|\textbf{t}_k-\textbf{t}_j\|^2}\rangle,\\
  &\langle\frac{\textbf{s}_i-\textbf{s}_j}{\|\textbf{s}_i-\textbf{s}_j\|^2},\frac{\textbf{s}_k-\textbf{s}_j}{\|\textbf{s}_k-\textbf{s}_j\|^2}\rangle),
\end{aligned}
\end{equation}
where $\ell_H$ denotes Huber loss, $\bl{t}_i=\Phi(\hat{\bl{x}}_i;\hat{\bl{w}})$ is the representation extracted by teacher recognizer, $\bl{s}_i=\phi(\bl{x}_i,\bl{m}_i;\{\bl{w}_e,\bl{w}_r\})$ is the discriminative representation output by the reformer. $\mu_{\bl{v}\in[\bl{t},\bl{s}]} = \frac{1}{|\bl{S}_2|}\sum_{(i,j) \in \bl{S}_2 } \| \bl{v}_i - \bl{v}_j \|^2$ normalizes distances between teacher and student representations into the same scale, which enables relational structure transfer. $\bl{S}_2 = [(i,j) | 1 \leq i, j \leq n, i \neq j]$ and $\bl{S}_3=[(i,j,k)|1\leq i,j,k \leq n, i\neq j\neq k]$ are pairwise set and triplet set, respectively. $\langle\rangle$ denotes cosine angle. The reformer training loss is re-formulated as:
\begin{equation}\label{hard}
\mathcal{L}_{d}(\bl{w}_r)=\mathcal{L}_1+\alpha\mathcal{L}_2+\beta\mathcal{L}_3,
\end{equation}
where $\mc{L}_1 = \sum_{i=1}^n ||\bl{t}_i-\ell_0(\bl{s}_i)||$ measures one order structural relation. $\ell_0(\cdot)$ is a linear mapping to convert the dimension of reformer output by adding a $2048$-way linear layer on its top, which can facilitate the pretraining. The two factors $\alpha$ and $\beta$ are used for balancing the loss terms, and set as 0.01 and 0.02, respectively. As shown in  Fig.~\ref{fig:representations}, the reformed discriminative representations are effectively clustered according to identity and present clear separation between clustering of different identities, proving their identity discriminability. Thus, both encoder together with reformer plays an important role in representations for masked face recognition task,  where the representations keep consistent with different masks and strengthen identity clues. 

\subsection{Feature Classifier}
Feature classifier predicts a face identity from the reformed discriminative representation. It presents as a simple classification head, with a fully-connected layer and a softmax layer. The fully-connected layer uses 512-way to reduce the feature dimension and model parameters. We cascade feature classifier with the trained backbone and perform an end-to-end finetuning by minimizing the classification loss $\mathcal{L}_{c}$, which is defined as the cross-entropy loss between classifier output $p_{i}=\phi(\bl{x}_i,\bl{m}_i;\{\bl{w}_e,\bl{w}_r,\bl{w}_c\})$ and the groundtrue identity label $c_{i}$ on training samples:
\begin{equation}\label{Lcls}
  \mathcal{L}_{c}(\bl{w}_c)=-\frac{1}{n}\sum_{i=1}^{n}c_i\log(p_{i}).
\end{equation}
%where $p_{i}$ denotes the predicted possibility of the sample $\bl{x}_i$ belonging to identity $c_i$.

\subsection{Discussion}
\myPara{Relationship with other approaches}~
Our approach can be seen as the fusion of generative approach with encoder-decoder architecture~\cite{li2017generative,elyor2017cvpr} and discriminative approach focusing on knowledge transfer with two-stream framework~\cite{wonpyo2019rkd,zhang2021ekd}, which transforms the representations from masked and groundtrue faces into a discriminative feature space. It learns general face knowledge with generative representations via inpainting like masked image modeling~\cite{he2022masked,xie2022simmim} but focuses on more fine-grained inpainting where the input is masked face instead of complete one. Thus, generative representations can evaluate the relationship between masks and masked faces. Moreover, it converts generative representations into discriminative ones using a reformer and a pretrained face recognizer, where pairwise and triplet knowledge like~\cite{schroff2015facenet,song2019iccv,li2020mm,boutros2022self} are transferred to facilitate identity recovery, rather than mean squared error in MaskInv~\cite{huber2021fg} and cosine distance in CSN~\cite{zhao2022spl}. Specially, our approach is beyond learning two cascaded ``vanilla'' networks which is hard to ensure their roles, and our main novelty is the greedy module-wise pretraining that combines the advantages of generative and discriminative representations by: 1) generative encoder that is finetuned via reconstruction to ensure its role in mask-robust representations, and 2) discriminative reformer that is trained via distillation to ensure its role in identity-robust representations.

\myPara{Network training} Due to greater complexity of masked face recognition and different learning objectives between generative encoder and discriminative reformer, training all modules altogether is hard to converge. Thus our network training includes finetuning generative encoder, learning discriminative reformer via distillation and finetuning feature classifier in a progressive manner. The main training cost comes from the learning of discriminative reformer and is similar to the training of general face recognition models~\cite{cao2018vggface2,deng2019arcface} even our entire network is larger.

\section{Experiments}
To verify the effectiveness of our generative-to-discriminative representation approach (\textbf{G2D}), we conduct experiments on both synthesized and realistic masked face datasets to provide comprehensive evaluations. 

\myPara{Datasets~}We use Celeb-A~\cite{Liu2015CelebA} for generating synthetic training data, LFW~\cite{LFWTech} for synthetic masked face evaluation, and RMFD~\cite{RMFD}  and MLFW~\cite{wang2022ccbr} for real-world masked face evaluation. {Celeb-A} consists of 202,599 face images covering 10,177 celebrities. Each face image is cropped, aligned by similarity transformation and then scaled to $256\times256$. We randomly split it into training set and validation set with the ratio of $6:1$.
{RMFD} consists of both synthetic and real-world masked faces with identity labels, covering various occlusion types and unconstrained scenarios. Our experiments only use the real-world masked face verification dataset, which contains 4,015 face images covering 426 subjects. The dataset is further organized to get 6,914 masked-unmasked pairs, including 3,457 positive and 3,457 negative pairs and serving as a valuable benchmark for cross-quality validation. 
MLFW is a relatively more difficult database to evaluate the performance of masked face verification. The dataset maintains the data size and the face verification protocol of LFW, considers that two faces with the same identity wear different masks and two faces with different identities wear the same mask, and emphasizes both the large intra-class variance and the tiny inter-class variance simultaneously.

For self-supervised backbone training, we synthesized massive masked faces via MaskTheFace~\cite{anwar2020masked}. For an input face, it detects the keypoints, applies affine transformation to a randomly selected mask, overlays the original image, and perform post-processing to obtain natural masked face. More details are given in Appendix~\ref{app:synthesis}.

\myPara{Baselines~}We consider four kinds of baselines: I) four general face recognizers  (CenterLoss~\cite{wen2016discriminative} (CL), VGGFace~\cite{parkhi2015deep} (VGG), VGGFace2~\cite{cao2018vggface2} (VGG2) and ArcFace~\cite{deng2019arcface} (AF)), II) generative approaches that equip the four general face recognizers with four face inpainting approaches (GFC~\cite{li2017generative}, DeepFill~\cite{yu2018generative}, IDGAN~\cite{ge2020tcsvt} and ICT~\cite{wan2021high}) and replace masked faces with inpainted faces as input, III) finetuning-based masked face recognizers, and IV) models trained on masked faces from scratch. Baselines in kind III and Kind IV are discriminative approaches. Baselines in kind IV adopt DoDGAN~\cite{li2020mm} which first performs inpainting then learns a specialized recognizer with inpainted faces as input. To ensure fair comparisons, for each baseline, we use its published pretrained model to obtain the results and follow the same protocols for data preparation.

\myPara{Evaluation~}We evaluate masked face verification under two settings: 1)~MR-MP denoting masked reference against masked probe for evaluating over masked face pairs, and 2)~UMR-MP standing for unmasked reference against masked probe, which is closer to real-world gallery-probe scenario. The evaluation is measured with 8 metrics, including verification accuracy (ACC), equal error rate (EER), Fisher discriminant ratio (FDR), false match rate (FMR), false non-match rate (FNMR), the lowest FNMR for a FMR $\leq1.0\%$ (FMR100), the lowest FNMR for a FMR $\leq0.1\%$ (FMR1000), and the average value calculated based on FMR100\_Th and FMR1000\_Th thresholds (AVG). The last 5 metrics are also used in~\cite{huber2021fg}.
% maio2020tpami

% ~\cite{paszke2019pytorch},~\cite{Kingma2014Adam}
\myPara{Implementation details~}The experiments are implemented on Pytorch. To get facial masks, we perform simple segmentation based on Grabcut~\cite{rother2004tog} automatically initialized the seeds with classical image features like colors and shapes. For generative encoder, we finetune ICT inpainting network with a batch size of $16$ using Adam optimizer, where learning rate is $10^{-5}$ and $\beta_1=0.5, \beta_2=0.9$. For discriminative reformer, we employ pretrained VGGFace2~\cite{cao2018vggface2} as teacher since its input size is the same to generative encoder. All models are trained with a batch size of $64$ and SGD optimizer. The initial learning rate is $0.1$ and decreases to $0.5$ times every $16$ epochs. The momentum and weight decay are set as $0.9$ and $5\times 10^{-4}$, respectively. 

\begin{figure}
  \centering
  \includegraphics[width=1.0\linewidth]{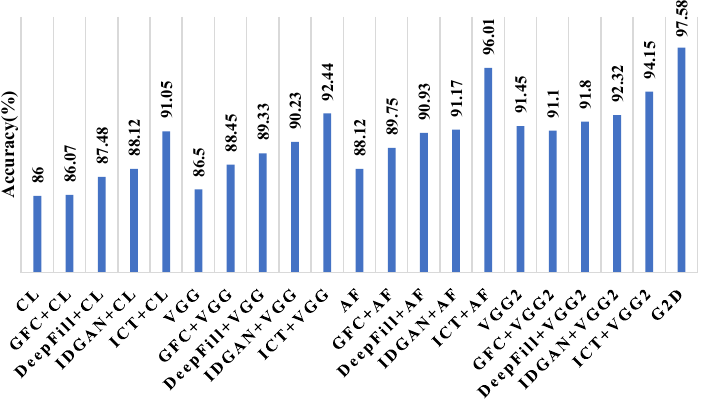}
  \caption{Evaluation on synthetic masked LFW. We report the accuracy of the proposed method (G2D), and make comparisons with combinations of general face recognizers (CenterLoss~\cite{wen2016discriminative} or CL, VGGFace~\cite{parkhi2015deep} or VGG, ArcFace~\cite{deng2019arcface} or AF, and VGGFace2~\cite{cao2018vggface2} or VGG2), and state-of-the-art generative face inpainting approaches (GFC~\cite{li2017generative}, DeepFill~\cite{yu2018generative}, IDGAN~\cite{ge2020tcsvt} and ICT~\cite{wan2021high}).}
  \label{fig:lfw_results}
\end{figure}

\begin{table*}[t]
\centering
\caption{Verification Performance on LFW synthetic masked faces under MR-MP and UMR-MP settings.}\label{tab:stats-lfw}
\resizebox{\textwidth}{!}{
\begin{tabular}{ccccccccccccc}
\toprule
&&&&&&&\multicolumn{3}{c}{FMR100\_Th}&\multicolumn{3}{c}{FMR1000\_Th}\\ \cmidrule{8-10} \cmidrule{11-13}
Setting & Model & ACC$\uparrow$ & EER$\downarrow$  & FDR$\uparrow$ & FMR100$\downarrow$ & FMR1000$\downarrow$ & FMR$\downarrow$ & FNMR$\downarrow$ & AVG$\downarrow$ & FMR$\downarrow$ & FNMR$\downarrow$ & AVG$\downarrow$ \\ \midrule

\multirow{15}{*}{MR-MP}&MFN~\cite{chen2018mobilefacenets} &81.53\%& 18.67\% & 1.53 & 61.23\%& 80.07\% & 2.63\% & 49.53\% & 26.08\% & 0.60\% & 69.13\% & 34.87\% \\ 
&ResNet50~\cite{he2015resnet} &85.85\%& 14.73\% & 2.03 & 46.17\%& 64.00\% & 1.77\% & 39.73\% & 20.75\% & \underline{0.07\%} & 65.07\% & 32.57\% \\
&ResNet100~\cite{he2015resnet} &92.27\%& 8.03\% & 3.42 & 21.53\%& 41.70\% & 2.53\% & 15.60\% & 9.07\% & 0.80\% & 24.27\% & 12.53\% \\
&DeepFill~\cite{yu2018generative}+CL~\cite{wen2016discriminative} &87.48\%& 13.43\% & 2.61 & 46.43\%& 63.20\% & 0.70\% & 50.13\% & 25.42\% & \underline{0.07\%} & 70.40\% & 35.23\% \\
&DeepFill~\cite{yu2018generative}+VGG~\cite{parkhi2015deep} &89.33\%& 11.00\% & 2.37 & 36.57\%& 57.43\% & 1.03\% & 36.57\% & 18.80\% & 0.10\% & 58.40\% & 29.25\% \\
&DeepFill~\cite{yu2018generative}+AF~\cite{deng2019arcface} &90.93\%& 9.27\% & 3.44 & 27.73\%& 54.67\% & 1.33\% & 25.13\% & 13.23\% & 0.10\% & 54.90\% & 27.50\% \\
&DeepFill~\cite{yu2018generative}+VGG2~\cite{cao2018vggface2} &91.80\%& 8.37\% & 4.42 & 27.40\%& 52.97\% & \underline{0.50\%} & 36.87\% & 18.68\% & \textbf{0.00\%} & 65.43\% & 32.72\% \\
&ICT~\cite{wan2021high}+CL~\cite{wen2016discriminative} &91.05\%& 8.98\% & 3.91 & 34.50\%& 73.13\% & 0.68\% & 37.08\% & 18.88\% & 0.14\% & 57.06\% & 28.60\% \\
&ICT+VGG~\cite{parkhi2015deep} &92.44\%& 7.59\% & 4.19 & 23.95\%& 45.86\% & 1.32\% & 21.37\% & 11.35\% & 0.17\% & 40.10\% & 20.13\% \\
&ICT+AF~\cite{deng2019arcface} &\underline{96.01\%}& \underline{3.97\%} & \underline{6.66} & \textbf{7.53\%} & \textbf{15.03\%} & \textbf{0.24\%} & 12.79\% & \textbf{6.51\%} & \textbf{0.00\%} & 43.49\% & 21.74\% \\
&ICT+VGG2~\cite{cao2018vggface2} &94.15\%& 6.00\% &  5.65 & 20.90\%& 38.36\% & 1.32\% & 18.89\% & 10.11\% & \textbf{0.00\%} & 49.39\% & 24.69\% \\
&MFN (SRT)~\cite{boutros2022self} &78.23\%& 22.30\% & 1.23 & 68.40\%& 85.10\% & 4.60\% & 46.07\% & 25.33\% & 1.03\% & 67.57\% & 34.30\% \\
&ResNet50 (SRT)~\cite{boutros2022self} &78.87\%& 21.70\% & 1.22 & 66.97\%& 79.17\% & 5.60\% & 44.27\% & 24.93\% & 0.90\% & 68.43\% & 34.67\% \\
&ResNet100 (SRT)~\cite{boutros2022self} &92.80\%& 7.63\% & 3.54 & 20.97\%& 35.37\% & 2.03\% & 14.77\% & \underline{8.40\%} & 0.67\% & 23.23\% & 11.95\% \\
&DoDGAN~\cite{li2020mm}&95.44\%& 6.12\% & 5.60 & 22.45\% & 58.97\% & 34.93\% & \textbf{0.46\%} & 17.70\% & 10.20\% & \textbf{3.52\%} & \underline{6.86\%} \\
\cmidrule{2-13}
&\textbf{Our G2D} &\textbf{97.58\%}& \textbf{3.27\%} & \textbf{7.01} & \underline{10.74\%}& \underline{33.44\%} & 20.94\% & \underline{5.83\%} & 13.39\% & 6.40\% & \underline{3.65\%} & \textbf{5.02\%} \\

\midrule

\multirow{15}{*}{UMR-MP}&MFN~\cite{chen2018mobilefacenets} &90.28\%& 9.87\% & 3.17 & 33.40\%& 49.23\% & 0.73\% & 37.90\% & 19.32\% & 0.07\% & 62.00\% & 31.03\% \\ 
&ResNet50~\cite{he2015resnet} &88.83\%& 11.70\% & 2.79 & 27.37\%& 51.70\% & 0.40\% & 33.67\% & 17.03\% & \underline{0.03\%} & 57.90\% & 28.97\% \\
&DeepFill \cite{yu2018generative}+CL \cite{wen2016discriminative} &90.22\%& 7.53\% & 4.69 & 23.87\%& 48.23\% & 0.40\% & 31.60\% & 16.00\% & 0.10\% & 52.90\% & 26.50\% \\
&DeepFill~\cite{yu2018generative}+VGG~\cite{parkhi2015deep} &86.90\%& 6.63\% & 3.53 & 21.13\%& 43.30\% & 0.87\% & 22.47\% & 11.67\% & 0.13\% & 42.27\% & 21.20\% \\
&DeepFill~\cite{yu2018generative}+AF~\cite{deng2019arcface} &93.28\%& 10.63\% & 3.05 & 30.67\%& 50.90\% & 0.43\% & 39.80\% & 20.12\% & \textbf{0.00\%} & 73.47\% & 36.73\% \\
&DeepFill~\cite{yu2018generative}+VGG2~\cite{cao2018vggface2} &92.65\%& 5.70\% & 5.96 & 18.67\%& 37.67\% & 0.30\% & 29.57\% & 14.93\% & \textbf{0.00\%} & 62.70\% & 31.35\% \\
&ICT~\cite{wan2021high}+CL~\cite{wen2016discriminative} &91.73\%& 8.33\% & 4.50 & 30.15\%& 59.50\% & 0.54\% & 34.72\% & 17.63\% & 0.13\% & 54.18\% & 27.16\% \\
&ICT+VGG~\cite{parkhi2015deep} &92.81\%& 7.26\% & 4.55 & 22.66\%& 43.60\% & 0.40\% & 29.55\% & 14.97\% & 0.07\% & 53.75\% & 26.91\% \\
&ICT+AF~\cite{deng2019arcface} &93.28\%& 7.36\%  & 4.32& 17.48\%& \underline{26.99\%} & \underline{0.03\%} & 34.55\% & 17.29\% & \textbf{0.00\%} & 75.09\% & 37.55\%  \\
&ICT+VGG2~\cite{cao2018vggface2} &94.99\%& 5.21\% &  \underline{6.41} & \underline{17.04\%}& 48.13\% & 0.91\% & 18.25\% & 9.58\% & 0.07\% & \underline{50.99\%} & \underline{25.53\%} \\
&MFN (SRT) \cite{boutros2022self} &87.97\%& 12.30\% & 2.65 & 40.53\%& 59.47\% & 0.23\% & 55.13\% & 27.68\% & \textbf{0.00\%} & 82.50\% & 41.25\% \\
&ResNet50 (SRT)~\cite{boutros2022self} &82.90\%& 17.70\% & 1.73 & 48.23\%& 65.27\% & \textbf{0.00\%} & 94.77\% & 47.38\% & \textbf{0.00\%} & 99.97\% & 49.98\% \\
&DoDGAN~\cite{li2020mm} &\underline{94.32\%}& \underline{5.02\%} & 5.46 & 19.41\% & 73.52\% & 4.28\% & \underline{8.92\%} & \underline{6.55\%} & 0.42\% & 51.50\% & 25.96\% \\
\cmidrule{2-13}
&\textbf{Our G2D} &\textbf{97.75\%} & \textbf{3.05\%} & \textbf{8.02}& \textbf{8.93\%}& \textbf{22.55\%} & 2.14\% & \textbf{2.67\%} & \textbf{2.41\%} & 0.17\% & \textbf{13.65\%} & \textbf{6.96\%}  \\\bottomrule

\end{tabular}}
\end{table*}

\subsection{Evaluation on Synthetic Masked Faces}
We report the performance on synthetic masked faces. Similar to training data, we generate synthetic masked faces using images from LFW for a comprehensive evaluation, achieving 3,000 positive pairs with the same identities and 3,000 impostor pairs with different identities.

\myPara{Comparison to baselines in kind I and kind II~}In this experiment, all recognizers and composite models extract features and then computes the cosine similarities for all the 6,000 face pairs. The accuracy is the percentage of correct predictions, where the threshold is decided as the one with the highest accuracy. 
The results are reported in Fig.~\ref{fig:lfw_results}. Three main conclusions can be drawn. First, diverse masks result in evident accuracy drop, which is in accord with previous research findings~\cite{ngan2020a, ngan2020b}. Second, generative face inpainting sometimes are not always able to fill the gap. We notice that the combination of VGG2 and GFC achieves even lower accuracy than VGG2 alone, suggesting that the inpainting process may play a negative role if it cannot be regularized properly. We suspect it is due to the interference of similar mask patterns and poor robustness of inpainting model. Third, in the face inpainting plus recognition paradigm, adoption of the inpainting method do make a difference to the performance of the composite model. {Moreover, on synthetic masked LFW, IDGAN delivers a 96.53\% accuracy under 48$\times$48 masks~\cite{ge2020tcsvt} when our G2D achieves 97.58\% even under more complex masks.} Finally, our G2D outperforms all combinations, proving the effectiveness of our approach.

\myPara{Comparison to baselines in kind III and kind IV~}Then, we employ the combinations of two inpainting approaches, DeepFill and ICT, with the four recognizers, together with two recently-proposed masked face recognition models, DoDGAN~\cite{li2020mm} and Self-Restrained Loss (SRT)~\cite{boutros2022self}, for more quantitative comparisons. Here, we do not adopt IDGAN, since it shares the same backbone with DeepFill and trained with full identity supervision. We intend to focus more on the efficacy of self-supervised representation learning. Tab.~\ref{tab:stats-lfw} presents the results under UMR-MP and MR-MP settings. For SRT, the performance of both baselines (ResNet50~\cite{he2015resnet} and MobileFaceNet~\cite{chen2018mobilefacenets}) and those along with an extra module trained with SRT loss are reported.

As shown in Tab.~\ref{tab:stats-lfw}, for SRT which finetunes existing deep recognizers with an extra module on top, the original baselines, instead of the refined ones, show better performance. This suggest that, although these solutions can recover some performance on masked samples, the generalization ability of deep models can be easily suffered. Similarly, the recent work DoDGAN~\cite{li2020mm} experienced an evident drop on cross-quality evaluation. In essence, these approaches do not appropriately handle the distribution divergence between masked and non-masked samples in the latent space. Our G2D achieves the highest accuracy on both MR-MP and UMR-MP settings.
Tab.~\ref{tab:stats-lfw} also reports the fisher discriminant ratio (FDR), which measures the distinguish ability of positive and negative pairs. Our approach shows better capacity to deal with the identity obscuring of masked faces.

\myPara{Analysis on FMR and FNMR results~}It is worth to note that, the approaches based on off-the-shelf face recognizers show lower false match rate (FMR). It suggests that they tend to predict more positive pairs (which share the same identity) as negative, while prediction over negative pairs is less affected.
To the contrary, our G2D shows evident superiority in false non-match rate (FNMR). This reveals a basic difference in our motivation. When occlusions occurs, for general face recognizers, the main challenge is the invalidation of pre-existing intra-class characteristics. Our approach, differently, teaches the model to doubt, and re-calibrate. It is also worth to note that, our model presents lower average values of FMR and FNMR, especially under UMR-MP setting. It suggests our proposed G2D achieves a better balance between FMR and FNMR, in another word, a better generalization over unmasked and masked faces.
\begin{table*}[t]
\centering
\caption{Performance on RMFD realistic masked faces under UMR-MP setting.}\label{tab:stats-whn}
\resizebox{\textwidth}{!}{\begin{tabular}{cccccccccccc}
\toprule
&&&&&&\multicolumn{3}{c}{FMR100\_Th}&\multicolumn{3}{c}{FMR1000\_Th}\\ \cmidrule{7-9} \cmidrule{10-12}
Model & ACC$\uparrow$ & EER$\downarrow$  & FDR$\uparrow$ & FMR100$\downarrow$ & FMR1000$\downarrow$ & FMR$\downarrow$ & FNMR$\downarrow$ & AVG$\downarrow$ & FMR$\downarrow$ & FNMR$\downarrow$ & AVG$\downarrow$ \\ \midrule
MFN~\cite{chen2018mobilefacenets} &69.90\%& 30.16\% & 0.49 & 88.69\%& 95.70\% & 1.01\% & 88.69\% & 44.85\% & 0.09\% & 95.70\% & 47.89\% \\
ResNet50~\cite{he2015resnet} &71.75\%& 28.44\% & 0.65 & 81.65\%& 94.32\% & 1.01\% & 81.65\% & 41.33\% & 0.09\% & 94.32\% & \underline{47.20\%} \\
MFN (SRT)~\cite{boutros2022self} &69.25\%& 31.28\% & 0.45 & 88.83\%& 97.09\% & \underline{0.09\%} & 97.14\% & 48.62\% & \underline{0.03}\% & 99.22\% & 49.63\% \\
ResNet50(SRT)~\cite{boutros2022self} &65.92\%& 34.76\% & 0.35 & 87.80\%& 96.88\% & \textbf{0.00\%} & 100.00\% & 50.00\% & \textbf{0.00}\% & 100.00\% & 50.00\% \\
ArcFace~\cite{deng2019arcface} & 72.35\% & \underline{27.71\%} & \underline{0.68} & \underline{81.59\%}& \underline{93.48\%} & 0.99\% & \underline{81.59\%} & \underline{41.29\%} & 0.09\% & \underline{93.48\%} & 46.78\%  \\
VGGFace2~\cite{cao2018vggface2} & 72.22\% & 27.91\% & 0.60 & 88.08\%& 98.67\% & 0.99\% & 88.08\% & 44.53\% & 0.09\% & 98.67\% & 49.38\% \\
DoDGAN~\cite{li2020mm} &\underline{72.55\%}& 28.26\%  & 0.54 & 83.12\%& 95.24\% & 1.01\% & 83.12\% & 42.07\% & 0.09\% & 95.24\% & 47.66\% \\
\midrule
\textbf{Our G2D} &\textbf{79.18\%}& \textbf{21.64\%}  & \textbf{1.31} & \textbf{72.18\%}& \textbf{86.89\%} & 0.99\% & \textbf{72.18\%} & \textbf{36.58\%} & 0.09\% & \textbf{86.89\%} & \textbf{43.49\%} \\
\bottomrule
\end{tabular}}
\end{table*}

\begin{figure}[!ht]
  \centering
  \setlength{\abovecaptionskip}{0.cm}
  \includegraphics[width=1.0\linewidth]{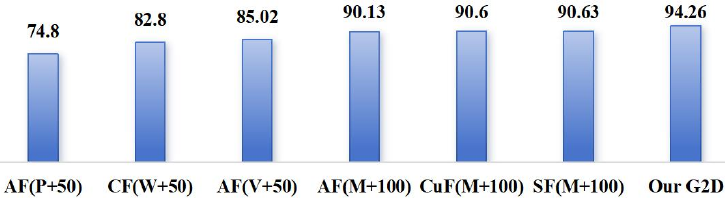}
  \caption{Verificaiton accuracy (\%) on MLFW~\cite{wang2022ccbr}. AF: ArcFace~\cite{deng2019arcface}, CF: CosFace~\cite{wang2018cvpr}, CuF: CurricularFace~\cite{huang2020cvpr}, SF: SFace~\cite{zhong2021tip}. P: Private-Asia, W: WebFace, V: VGGFace2, M: MS1MV2. 50 means ResNet50 and 100 means ResNet100.}
  \label{fig:mlfw_results}
\end{figure}

\subsection{Evaluation on Realistic Masked Faces}
We then evaluate on RMFD~\cite{RMFD}, where realistic masked faces have various mask types and complicated photographic conditions. We use 6,992 sample pairs to examine model performance and present comparison results in. Tab.~\ref{tab:stats-whn}. We can find that general models trained on normal faces all exhibit more violent drop on accuracy. For example, VGGFace2 achieves a 91.45\% accuracy on synthesized masked faces, while only gets a 72.22\% accuracy on realistic masked faces. The results prove the difficulty of the dataset. Our G2D achieves the highest accuracy of 79.18\%, which proves it capable of adapting to masked face recognition in the wild. The models with SRT show extreme imbalance between FMR and FNMR, which suggests they get overfitting to the masked face recognition scenarios while almost completely sacrificing the discriminant over unmasked faces. Instead, our G2D show better capacity to connect unmasked and masked faces, which is rather valuable in realistic applications. The evaluation on MLFW (Fig.~\ref{fig:mlfw_results})
also shows that our G2D delivers the best accuracy.

\begin{figure}[t]
\centering
\includegraphics[width=1.0\linewidth]{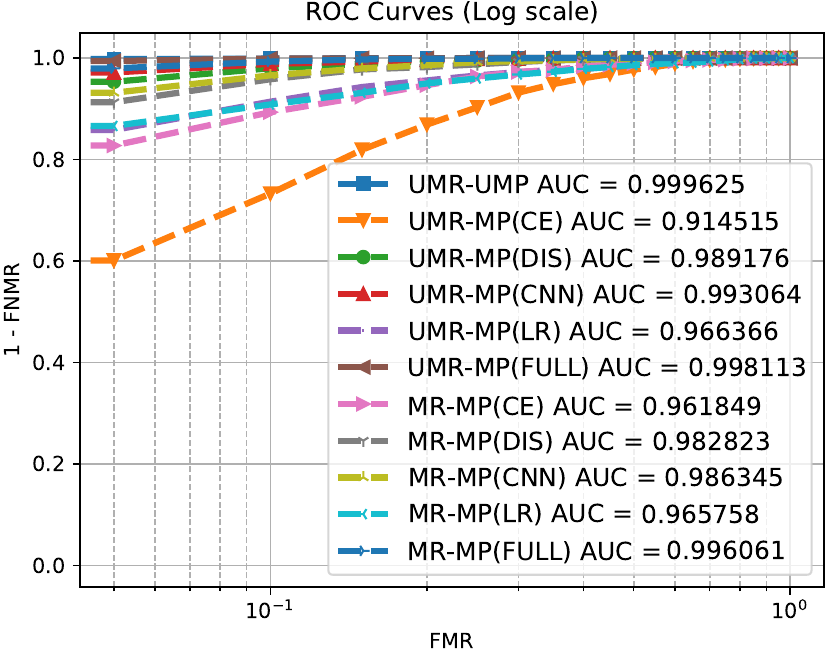}
\caption{The achieved log-scale ROC curves by G2D models trained with different losses. In each plot, the curves of UMR-MP cases are marked with a dashed line. The curves of MR-MP cases are marked with a dotted line. For each ROC curve, the area under the curve (AUC) is listed inside the plot.}
\label{fig:roc}
\end{figure}

\subsection{Ablation Studies}
\myPara{Generative encoder~}To evaluate our generative encoder design, we simulate the case when available information is reduced so that the model training cannot use the higher-level features, and compare with two variants: 1)~G2D(CNN) that uses a CNN-based inpainting network DeepFill as encoder and extracts generative representations from the layer before the decoding part of its second fine-grained network, and 2) G2D(LR) that removes convolutional layers from ICT upsampling network and takes appearance prior output as generative representations. From Tab.~\ref{tab:stats-lfw-g2d}, it is obvious that G2D outperforms G2D(CNN) due to better encoder, while the reduced input information leads G2D(LR) to overfitting and the learned representations poor generalization to different data domains. Fig.~\ref{fig:roc} provides their log-scale ROC curves, which also shows the similar conclusion, implying that Transformer-based generative encoder is more suitable for masked face recognition. We suppose that the representation space constructed by pretrained Transformer allows it to simulate and explore the distribution correlation among masked face, the corresponding mask and original face. 

\begin{table}[t]
\centering
\caption{Ablation study of G2D variants with different encoders and reformers under UMR-MP and MR-MP settings.}\label{tab:stats-lfw-g2d}
\resizebox{\linewidth}{!}{
\begin{tabular}{ccccccc}
\toprule
\multicolumn{7}{c}{\small\textbf{{Synthetic masked LFW}}}\\
\midrule
Setting & Model & ACC$\uparrow$ & EER$\downarrow$ & FDR$\uparrow$ & FMR100$\downarrow$ & FMR1000$\downarrow$ \\ \midrule
\multirow{5}{*}{UMR-MP}&G2D(CNN)&\underline{96.42\%} & \underline{3.60\%}& \underline{7.20}& \underline{14.20\%}& 46.10\% \\
&G2D(LR) &93.99\% & 6.54\% & 4.87& 17.98\%& \underline{38.77\%} \\
&G2D[CE] &83.50\%& 16.60\% & 1.94 & 69.83\% & 90.67\%  \\
&G2D[DIS] &95.25\%& 4.80\% & 6.04 & 18.30\% & 70.90\% \\
&{G2D} &\textbf{97.75\%} & \textbf{3.05\%} & \textbf{8.02}& \textbf{8.93\%}& \textbf{22.55\%} \\
\midrule

\multirow{5}{*}{MR-MP}
&G2D(CNN) &\underline{96.14\%}& \underline{5.77\%} & \underline{5.93} & \underline{18.40\%}& 55.63\% \\
&G2D(LR) &91.48\%& 9.96\% & 3.82 & 23.83\%& \underline{39.12\%} \\
&G2D[CE] &82.72\%& 10.40\% & 3.36 & 36.03\% & 69.37\% \\
&G2D[DIS] &93.53\%& 6.53\% & 5.49 & 24.37\% & 60.07\% \\
&{Full} &\textbf{97.58\%}& \textbf{3.27\%} & \textbf{7.01} & \textbf{11.74\%}& \textbf{38.44\%} \\\bottomrule
\end{tabular}}

\vspace{5pt}

\resizebox{\linewidth}{!}{
\begin{tabular}{ccccccc}
\toprule
\multicolumn{7}{c}{\small\textbf{{Realistic masked RMFD}}}\\
\midrule
Setting & Model & ACC$\uparrow$ & EER$\downarrow$ & FDR$\uparrow$ & FMR100$\downarrow$ & FMR1000$\downarrow$ \\ \midrule
\multirow{5}{*}{UMR-MP}&G2D(CNN) &73.26\%& 27.77\% & 0.61 & \underline{83.69\%}& \underline{93.27\%}  \\
&G2D(LR) &\underline{73.45\%}& \underline{27.02\%} & \underline{0.70} & 86.11\%& 95.29\%\\
&G2D[CE] &64.87\%& 35.37\% & 0.28 & 94.20\%& 99.02\% \\
&G2D[DIS] &70.80\%& 30.59\% & 0.48 & 87.21\%& 97.14\% \\
&{G2D} &\textbf{79.18\%}& \textbf{21.64\%} & \textbf{1.31} & \textbf{72.18\%}& \textbf{86.89\%} \\\bottomrule
\end{tabular}}
\end{table}

\myPara{Discriminative reformer~}First, we argue that discriminative reformer is very necessary due to poor identity discriminability of generative representations, \eg, the model achieves only a low accuracy of 57.10\% on synthetic masked LFW if discriminative reformer is discarded from the whole network. Then, we further check the learning process of discriminative reformer by comparing two models trained with different losses: 1) G2D[CE] trained with $\mc{L}_{c}$ only, and 2) G2D[DIS] trained with $\mc{L}_1$ only. The models trained with $\mc{L}_2$ and $\mc{L}_3$ can hardly converge, therefore the results are not presented. We report the results in Tab.~\ref{tab:stats-lfw-g2d} and Fig.~\ref{fig:roc}, which suggest that directly enforcing the model to approximate the hard identity label is less efficient. Thus, it is necessary to perform student learning supervised by a pretrained teacher whose features contain rich identity relationship~\cite{li2020mm}. A better teacher may lead to improved performance, \eg, we replace VGGFace2 with ArcFace as teacher where the inputs are resized into 112$\times$112, achieving a higher verification accuracy of 98.02\% on synthetic masked LFW than 97.58\% with VGGFace2 as teacher (Tab.~\ref{tab:stats-lfw}). 

\subsection{Further Analysis} 
\begin{table}[!ht]
  \centering
  \center
  \caption{Test accuracy (\%) on CPLFW~\cite{zheng2018cplfw}}.
   \small
    \begin{tabular}{cccccc}
    \hline
    CL &SphereFace & VGG2 & AF &MaskInv &G2D \\
    \hline
    77.48 & 81.40 & 84.00 & 92.08 &\textbf{92.86} &\underline{92.23}\\
    \hline
    \end{tabular}%
  \label{tab:res-cplfw}%
\end{table}%

\begin{figure}[!h]
\centering
{\includegraphics[width=0.47\linewidth]{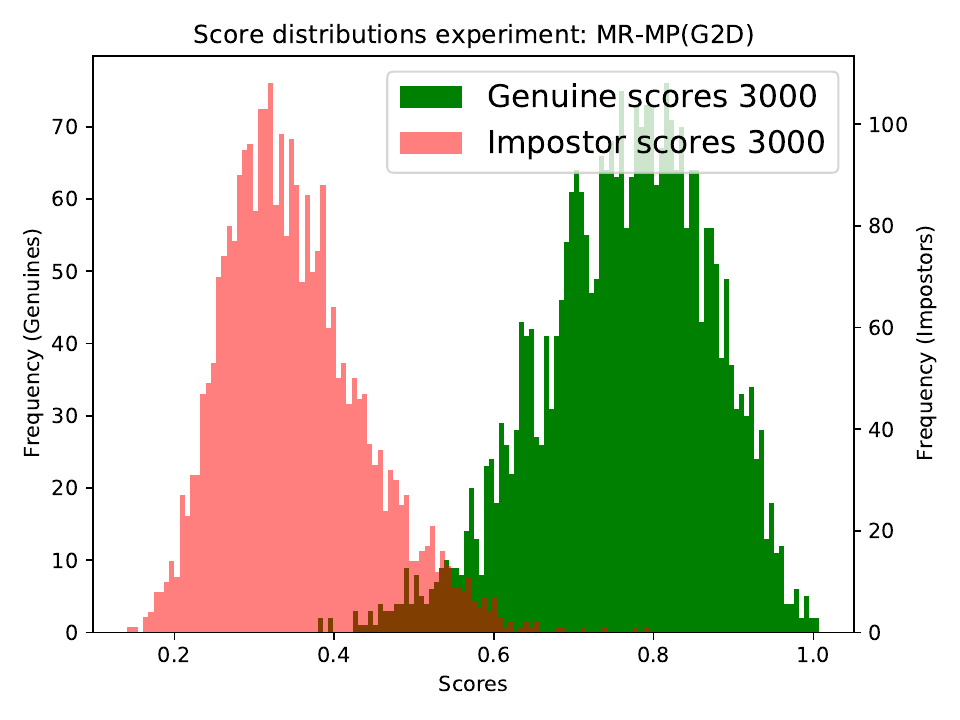}}
{\includegraphics[width=0.47\linewidth]{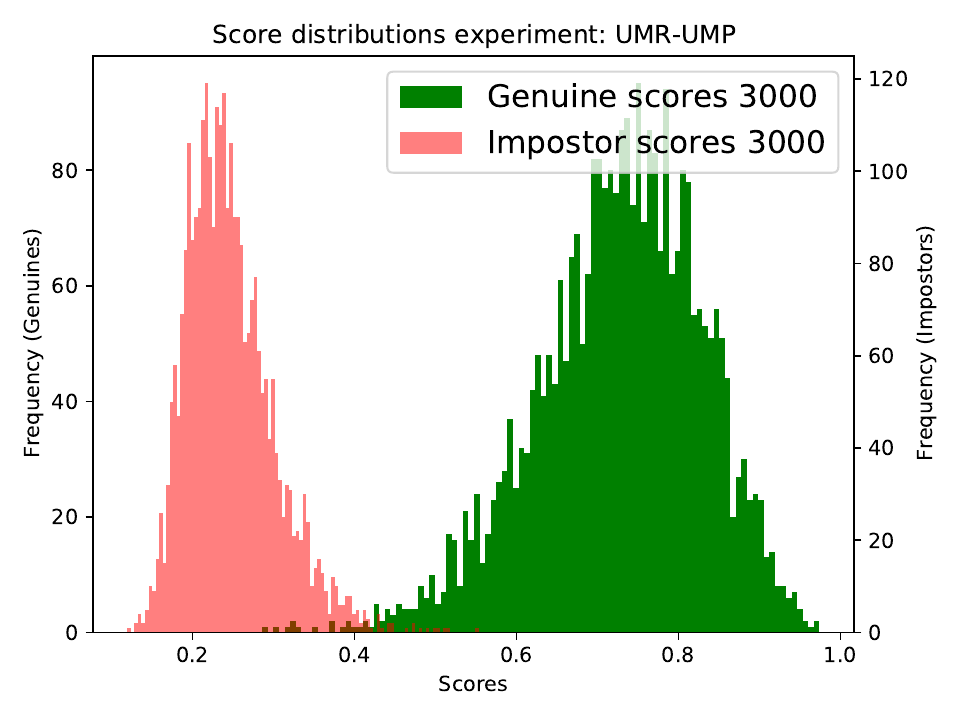}}
\hfil
\caption{Similarity score distributions of our G2D under MR-MP setting (left) and the ideal case under UMR-UMP setting (right). The positive and negative pairs are marked in green and red, respectively. Our G2D delivers small overlapping, which is close to the ideal case. More results are shown in Appendix~\ref{app:results}.}
\label{fig:score-distribution}
\end{figure}

\myPara{Evaluation on normal face recognition~}To evaluate the effect of our G2D on normal face recognition, we conduct an experiment on the normal face benchmark CPLFW~\cite{zheng2018cplfw} and report the results in Tab.~\ref{tab:res-cplfw}. We can find that our model achieves competitive performance, \eg, just lower than MaskInv~\cite{huber2021fg}. We suppose the main reasons include: 1) generative encoder that provides general and robust representations towards normal and masked faces, and 2) discriminative reformer that remains performance on normal faces by distilling on pretrained high-accuracy face recognizer.

\myPara{Representation discriminability~}Fig.~\ref{fig:representations} has showed that the reformed discriminative representations cluster the masked faces with the same identity together and present strong discriminability between different identities. We further conduct evaluation by using similarity score distributions on synthetic masked LFW and report the results in Fig.~\ref{fig:score-distribution}. We can find that our G2D delivers a small overlapping between positive and negative samples, which is close to the ideal case, demonstrating strong discriminability of our generative-to-dicriminative representations. 

\myPara{Inference efficiency~}Due to greater complexity of masked face recognition, our model has 178.5M parameters, larger than normal face recognition models (\eg, VGGFace2 and Arcface) who use Resnet50 as backbone and have 25.6M parameters. However, it is still efficient. We conduct efficiency analysis on a NVIDIA GeForce RTX 3090 GPU by performing inference on 100 masked faces with size of $256\times256$. The average inference time cost of a face image is 0.0428 seconds, leading to an inference speed of 23.35 FPS, implying the deployment feasibility in practical scenarios like urban governance. 

\section{Conclusion}
Masked face recognition has been gathering intensive attention over the past few years due to its real-world applications (\eg, fighting the COVID-19 pandemic). In this work, we propose to address masked face recognition by learning generative-to-discriminative representations. Our approach splits a unified network into three modules and learn them in a greedy module-wise pretraining way. Generative encoder and discriminative reformer are cascaded as the backbone to provide occlusion-robust and discriminative representations towards masked faces. The experiments are conducted on synthetic and realistic datasets to verify the effectiveness of our approach. In the future, we will design simultaneous training with synthetic and realistic datasets, and extend the framework to more vision tasks like occluded person ReID. 

\section*{Impact Statements}
This paper presents work whose goal is to advance the field of Machine Learning for Social Good, particularly addressing the challenge of masked face recognition. The proposed method would contribute positively to society by identifying masked faces and facilitating the development of Safety AI, \eg, improving urban governance and fighting  the COVID-19 pandemic. There are many other potential societal consequences of our work, none of which we feel must be specifically highlighted here.

% Acknowledgements should only appear in the accepted version.
%\section*{Acknowledgements}
\section*{Acknowledgements}
This work was supported by grants from the Pioneer R\&D Program of Zhejiang Province (2024C01024), and Open Research Project of the State Key Laboratory of Media Convergence and Communication, Communication University of China (SKLMCC2022KF004).
%\textbf{Do not} include acknowledgements in the initial version of
%the paper submitted for blind review.

% In the unusual situation where you want a paper to appear in the
% references without citing it in the main text, use \nocite
%\nocite{langley00}

\bibliography{main}
\bibliographystyle{icml2024}

%%%%%%%%%%%%%%%%%%%%%%%%%%%%%%%%%%%%%%%%%%%%%%%%%%%%%%%%%%%%%%%%%%%%%%%%%%%%%%%
%%%%%%%%%%%%%%%%%%%%%%%%%%%%%%%%%%%%%%%%%%%%%%%%%%%%%%%%%%%%%%%%%%%%%%%%%%%%%%%
% APPENDIX
\newpage
\appendix
\onecolumn

\section{The Synthesis of Masked Faces}\label{app:synthesis}
\begin{figure*}[!ht]
\centering
\includegraphics[width=0.7\linewidth]{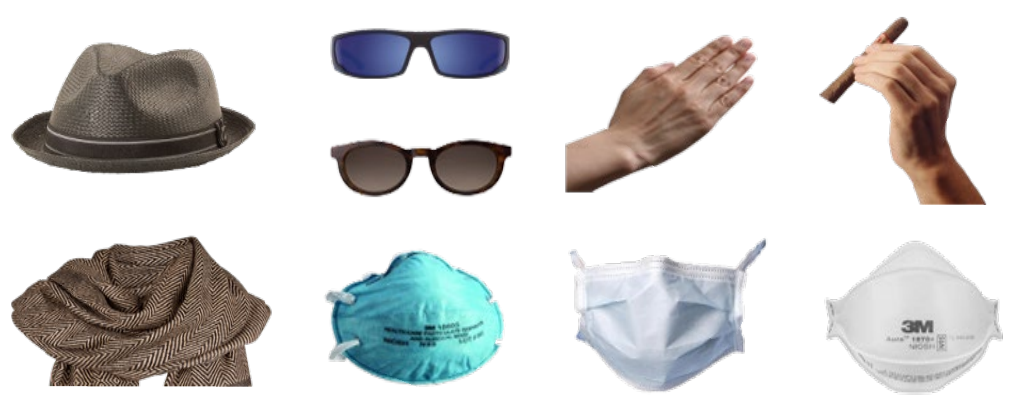}
\includegraphics[width=0.7\linewidth]{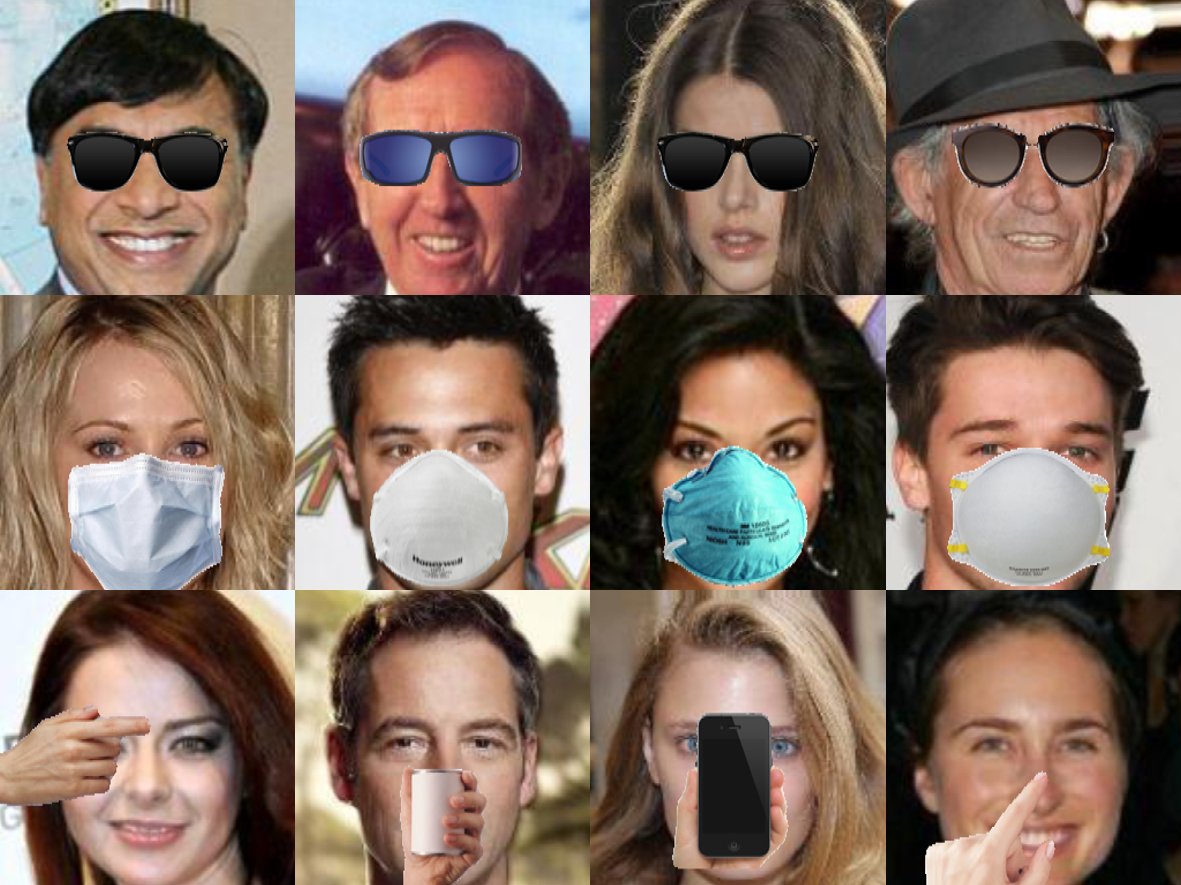}
\caption{Some examples of mask images (top) used for generating masked faces (bottom).}
\label{fig:mask-synthetic}
\end{figure*}

Our approach uses synthetic masked faces to train the models in a self-supervised manner. To this end, the masked faces are generated by synthesizing from normal faces. We take 202,599 normal facial images from Celeb-A dataset and synthesize massive masked facial images via pasting diverse mask patterns onto the images. To achieve that, we collected 45 transparent mask images (some examples are shown in Fig. \ref{fig:mask-synthetic}) online and resized them to cover an average of about 1/5 of the face. For a normal facial image, a random mask pattern is selected and simple alignment based on the facial landmarks is conducted to better simulate the realistic masked faces. Fig. \ref{fig:mask-synthetic} also presents some examples of the synthesized masked faces. To improve model generalizability, we further perform data augmentation by flipping and translation. 
%It detects keypoints and applies affine transformation to a randomly selected mask, overlays the original image, and perform post-processing to obtain natural masked faces.

\section{Representation Discriminability}\label{app:results}
\begin{figure*}[!ht]
\centering
\subfigure[MFN (SRT)]
{\includegraphics[width=0.25\linewidth]{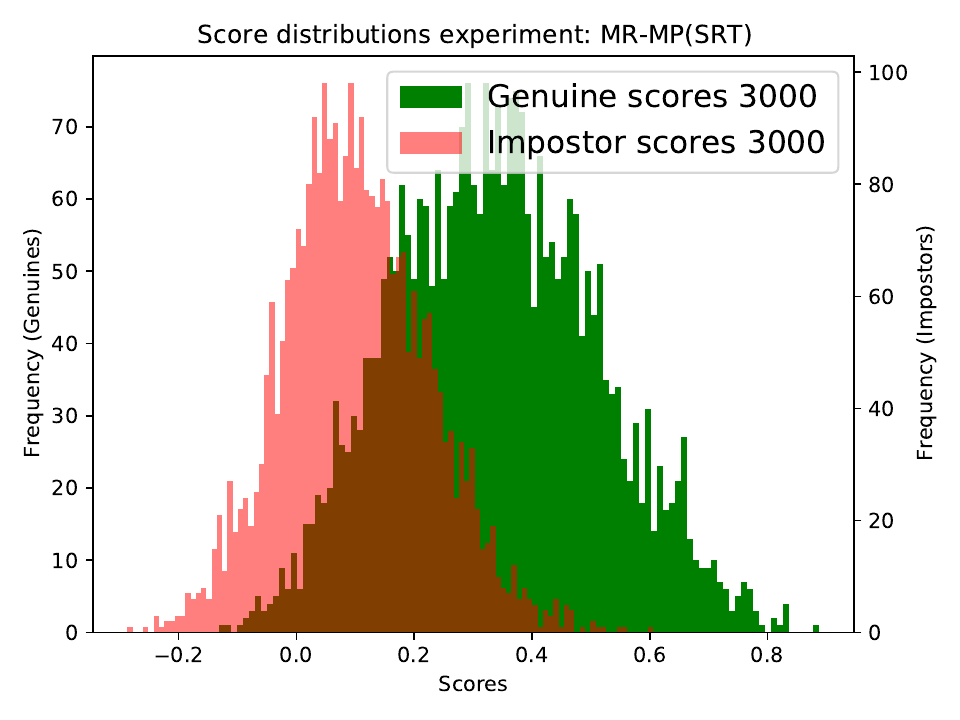}}
	\hfil
\subfigure[MFN]
{\includegraphics[width=0.25\linewidth]{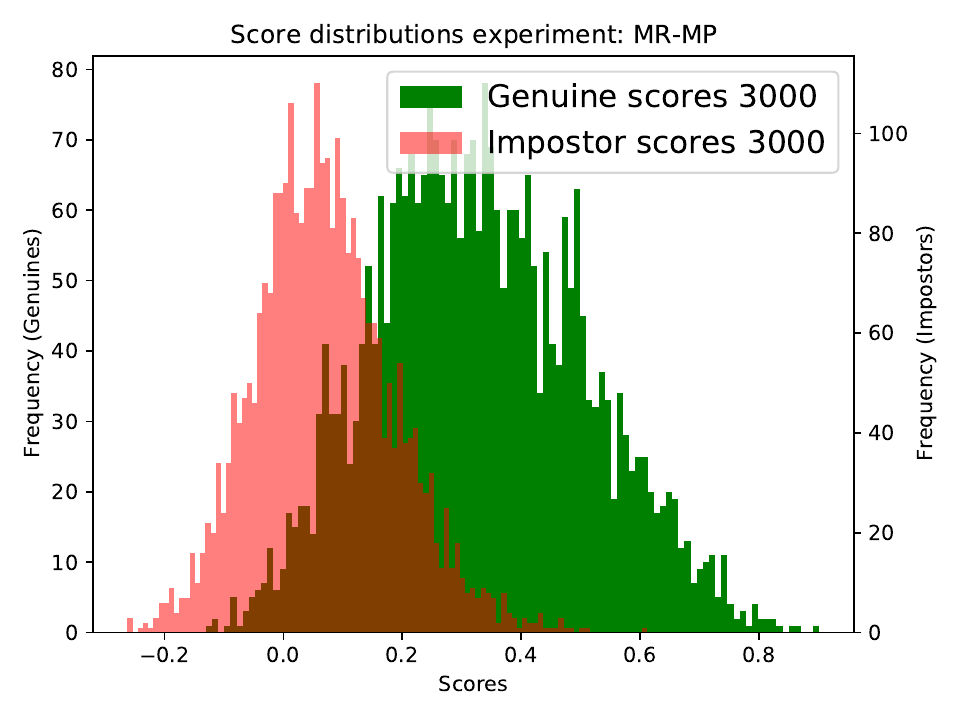}}
	\hfil
\subfigure[ResNet50 (SRT)]
{\includegraphics[width=0.25\linewidth]{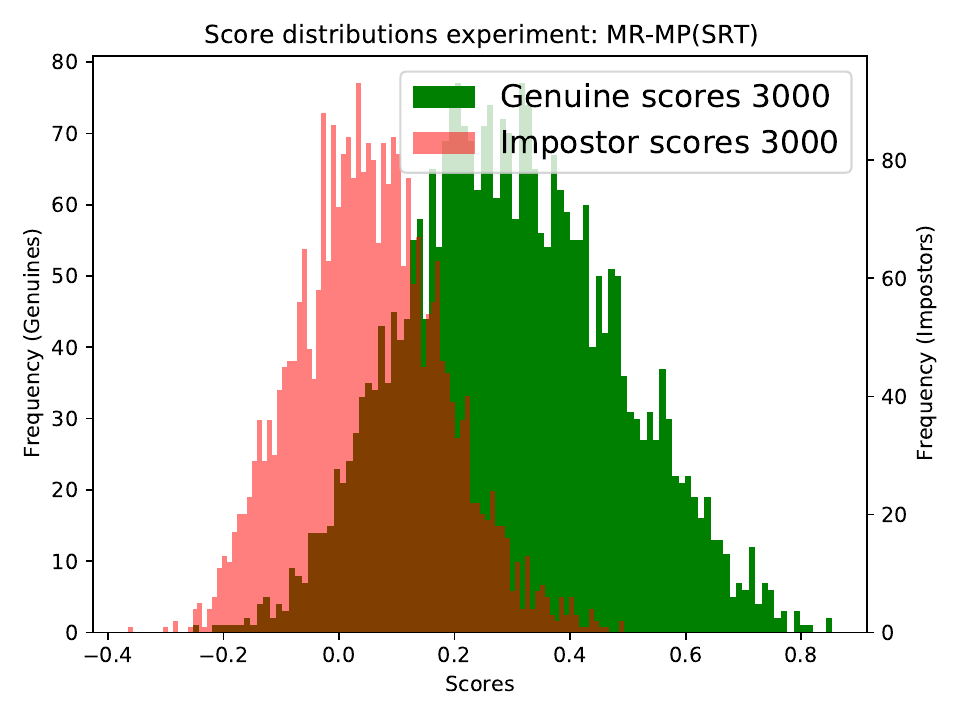}}
	\hfil
\subfigure[ResNet50]
{\includegraphics[width=0.25\linewidth]{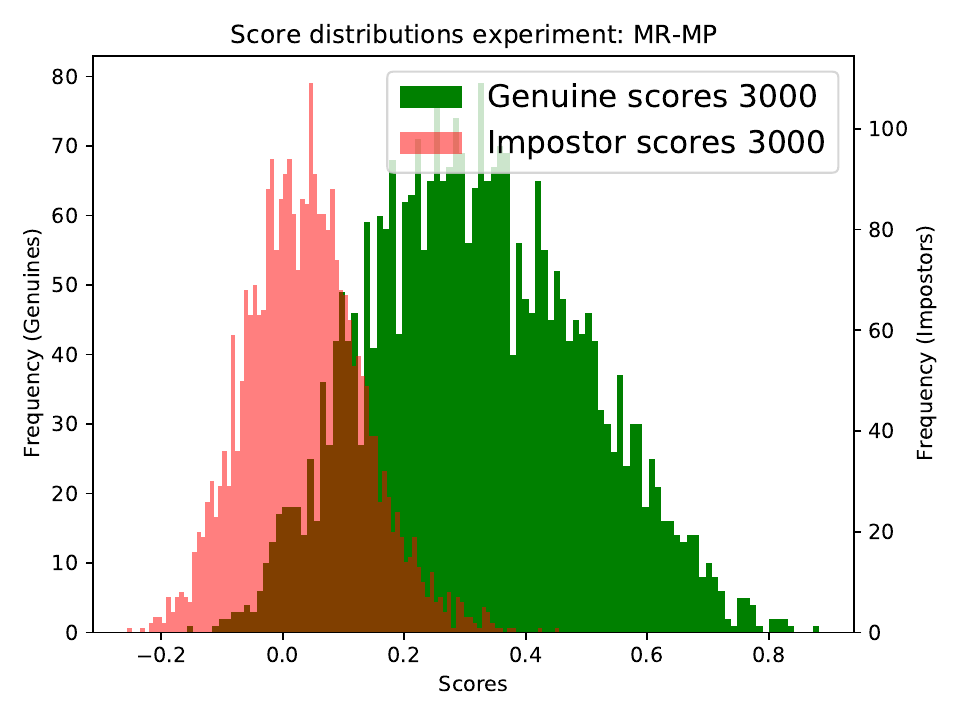}}
	\hfil
\subfigure[DeepFill+CL]
{\includegraphics[width=0.25\linewidth]{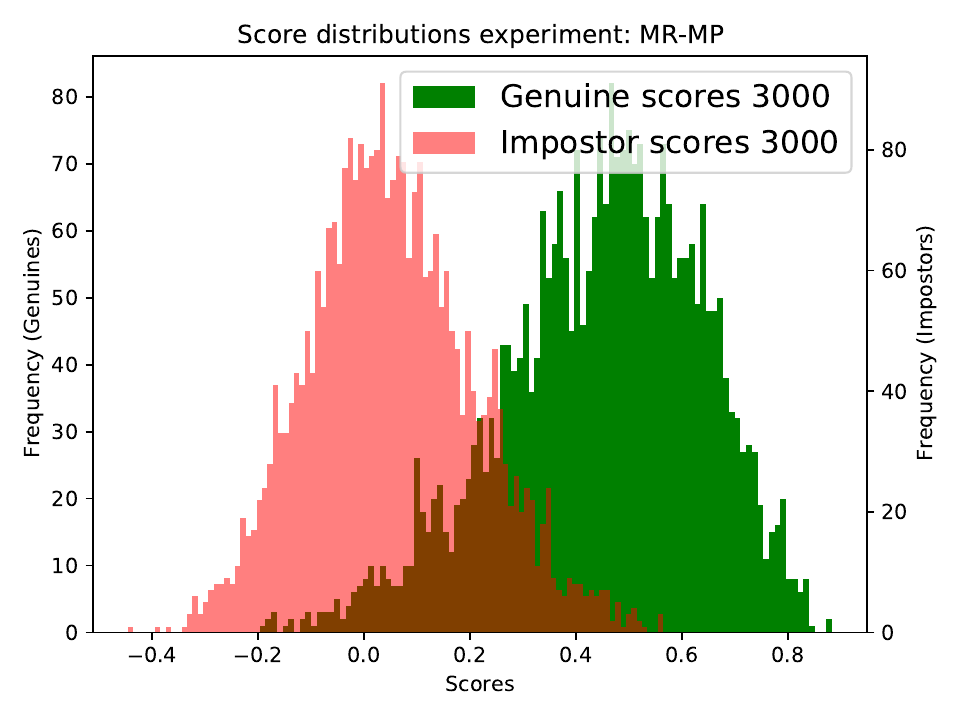}}
	\hfil
\subfigure[DeepFill+VGG]
{\includegraphics[width=0.25\linewidth]{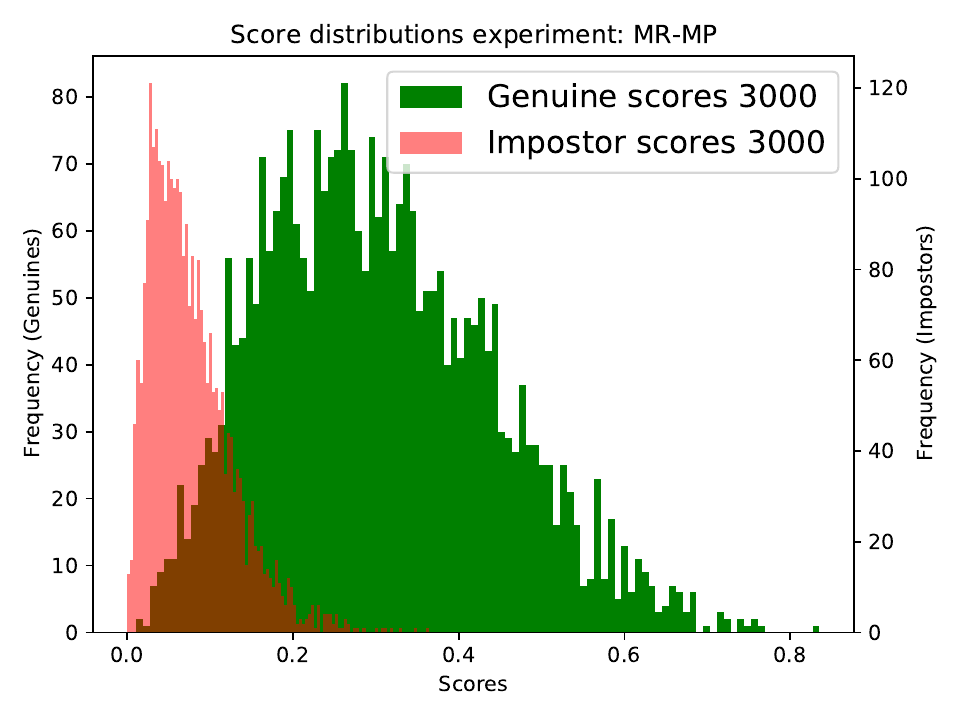}}
	\hfil
\subfigure[DeepFill+AF]
{\includegraphics[width=0.25\linewidth]{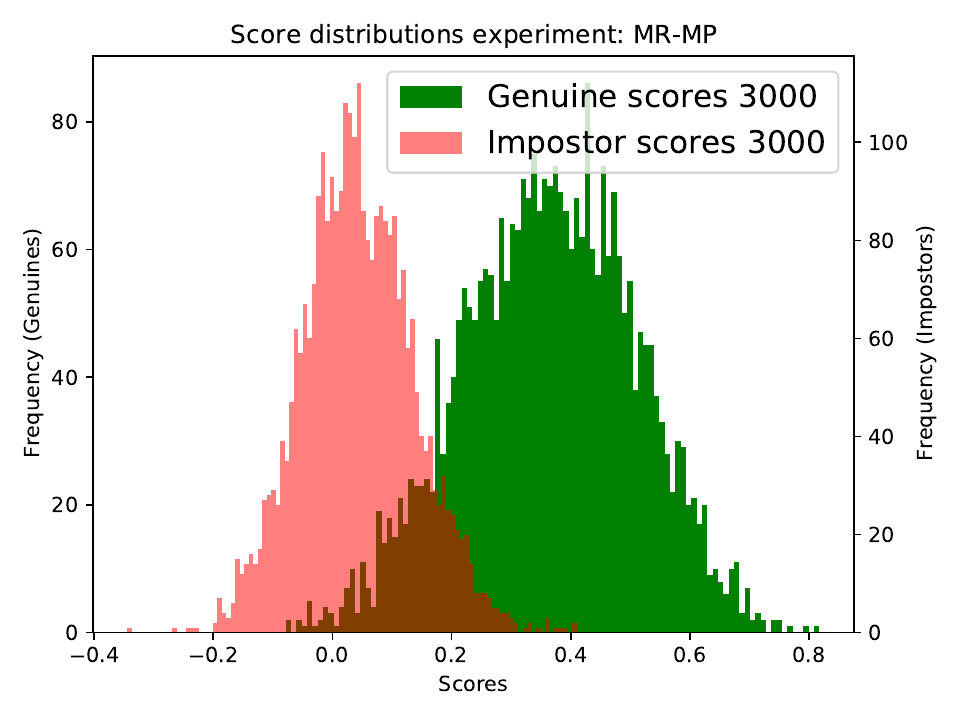}}
	\hfil
\subfigure[DeepFill+VGG2]
{\includegraphics[width=0.25\linewidth]{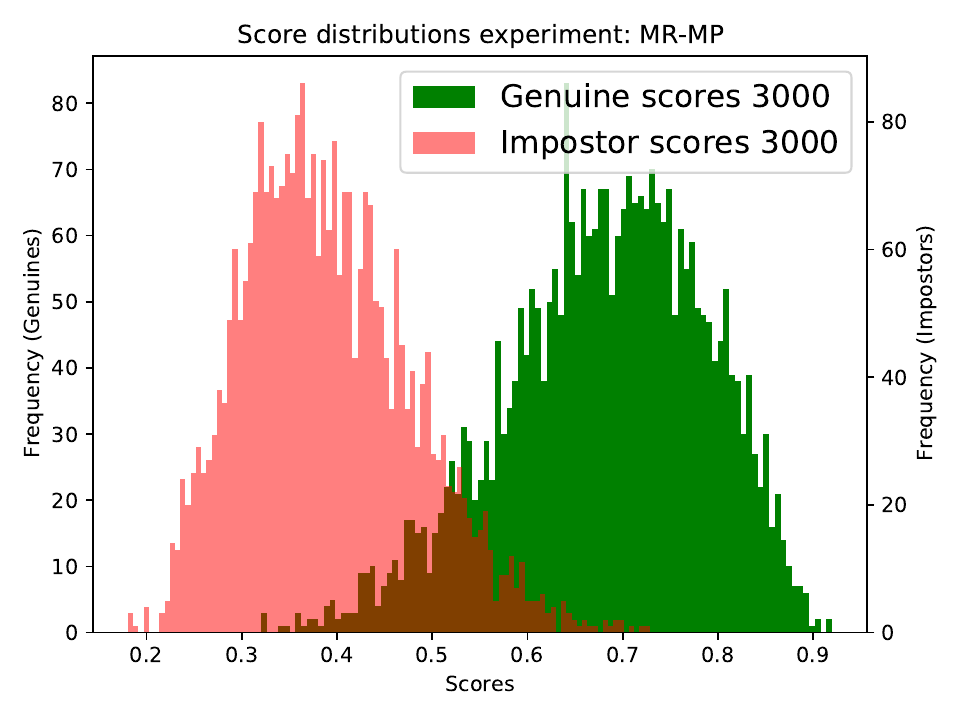}}
	\hfil
\subfigure[ICT+CL]
{\includegraphics[width=0.25\linewidth]{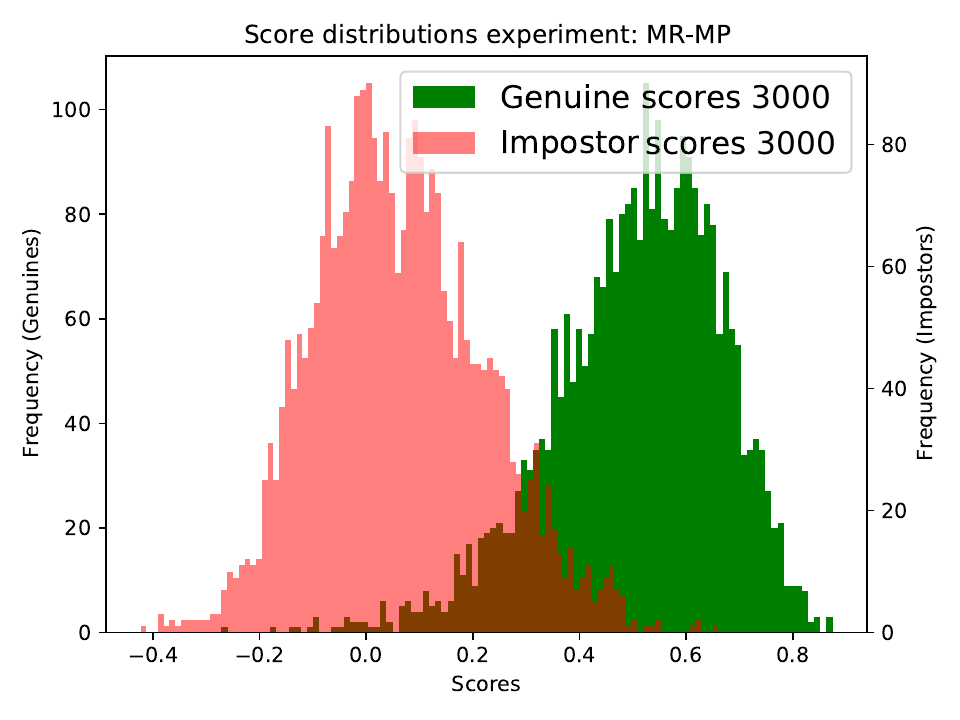}}
	\hfil
\subfigure[ICT+VGG]
{\includegraphics[width=0.25\linewidth]{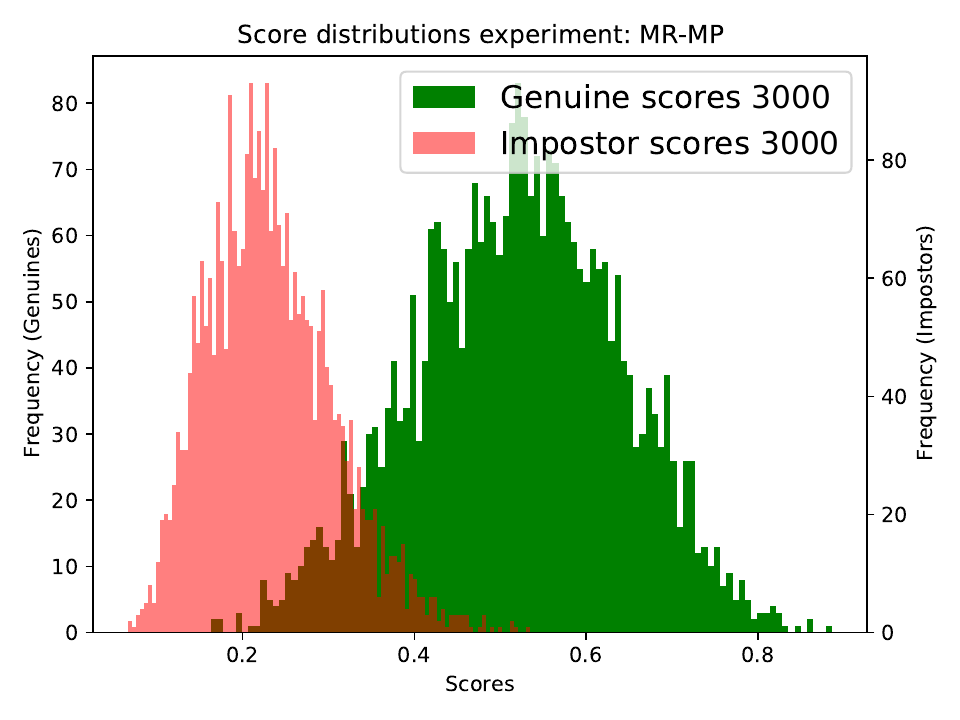}}
	\hfil
\subfigure[ICT+AF]
{\includegraphics[width=0.25\linewidth]{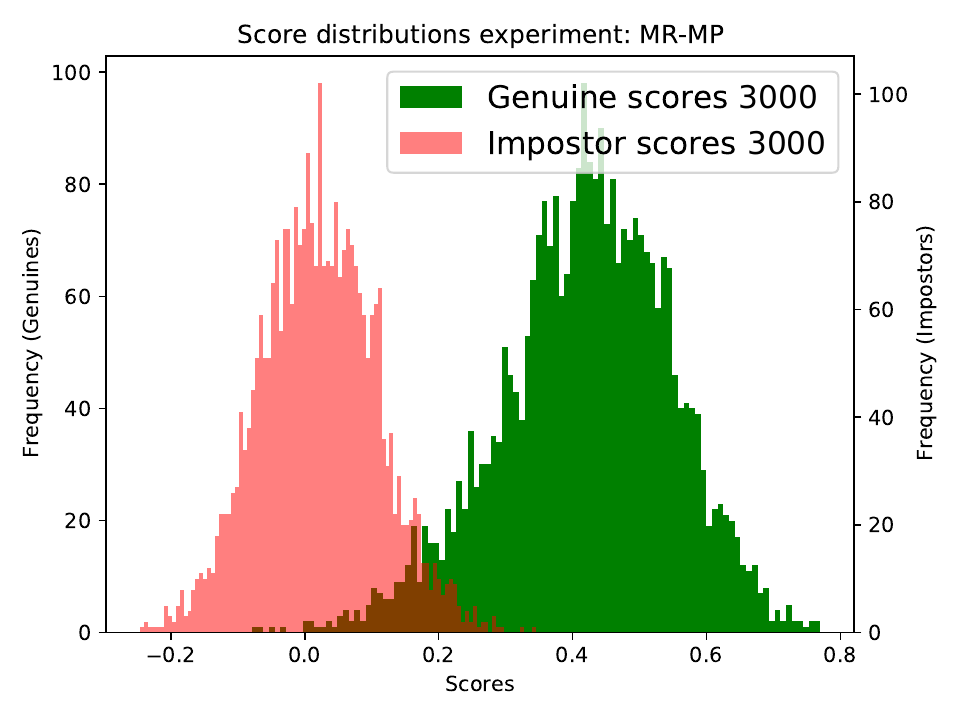}}
\hfil
\subfigure[ICT+VGG2]
{\includegraphics[width=0.25\linewidth]{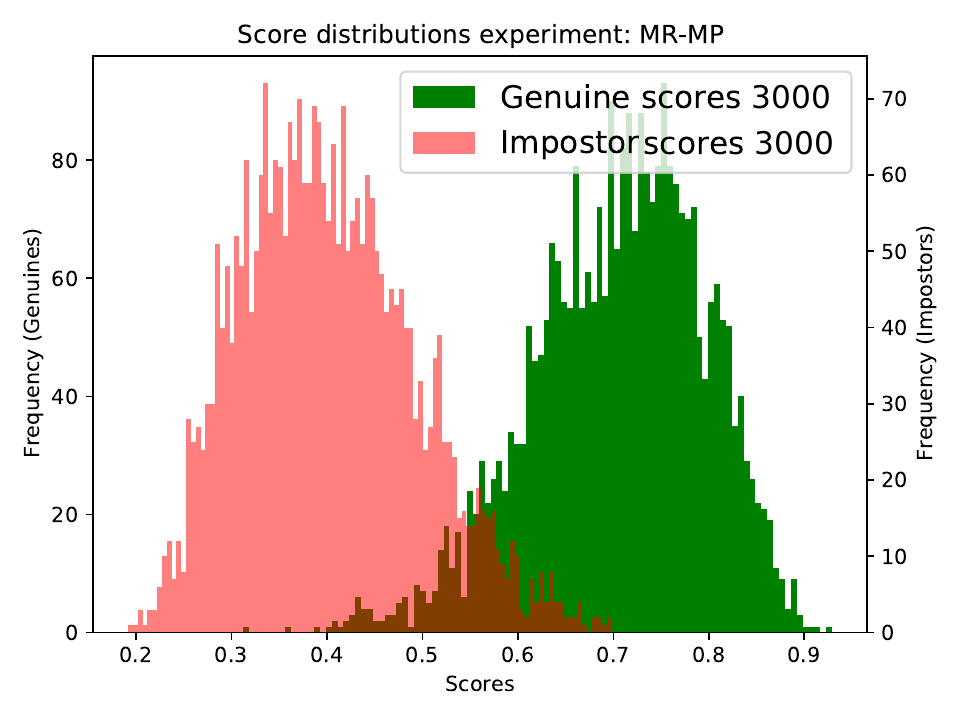}}
\hfil
\subfigure[Our G2D]
{\includegraphics[width=0.25\linewidth]{newplot/roc_log_cos_um_LFR/Distributions_MR-MP_G2D-T___lfw_LFR_Baseline_Distribution.pdf}}
\hfil
\subfigure[Ideal Case with UMR-UMP]
{\includegraphics[width=0.25\linewidth]{newplot/roc_log_cos_um_LFR/UMR-UMP_lfw_LFR_Baseline_Distribution.pdf}}
\hfil
\caption{The similarity score distributions achieved by different models under MR-MP setting. MR-MP briefs for masked/unmasked reference and masked/unmasked probes. The similarity score of the genuine pairs are in green color, and impostor pairs in red. Smaller overlapping suggest better discriminative ability. UMR-UMP refers to the circumstance where both probe and reference are unmasked, indicating the normal ideal case.}
\label{fig:distribution}
\end{figure*}

We can use similarity score distributions to evaluate the representation discriminability. Fig.~\ref{fig:distribution} reports the results achieved by different models on synthetic masked LFW under MR-MP setting. The scores of the genuine pairs are in green color, while the scores of the
impostor pairs (negative pairs) are presented in red color. Smaller overlapping
areas suggest a more distinct separation between pairs with
same and different identities. It illuminates that our G2D delivers smaller overlapping region than other models and is close to the ideal case (Fig.~\ref{fig:distribution} (n)), indicating that G2D can extract robust representations and provide stronger discriminative ability for masked faces. 

%%%%%%%%%%%%%%%%%%%%%%%%%%%%%%%%%%%%%%%%%%%%%%%%%%%%%%%%%%%%%%%%%%%%%%%%%%%%%%%
%%%%%%%%%%%%%%%%%%%%%%%%%%%%%%%%%%%%%%%%%%%%%%%%%%%%%%%%%%%%%%%%%%%%%%%%%%%%%%%
%%%%%%%%%%%%%%%%%%%%%%%%%%%%%%%%%%%%%%%%%%%%%%%%%%%%%%%%%%%%%%%%%%%%%%%%%%%%%%%
%%%%%%%%%%%%%%%%%%%%%%%%%%%%%%%%%%%%%%%%%%%%%%%%%%%%%%%%%%%%%%%%%%%%%%%%%%%%%%%

\end{document}